\pdfoutput=1

\documentclass[11pt]{article}

\usepackage[]{latex/acl}

\usepackage{times}
\usepackage{latexsym}

\usepackage[T1]{fontenc}

\usepackage[utf8]{inputenc}

\usepackage{microtype}

\usepackage{inconsolata}

\usepackage{times}
\usepackage{latexsym}
\usepackage{color}
\usepackage{fontawesome}
\usepackage[T1]{fontenc}

\usepackage[utf8]{inputenc}

\usepackage{microtype}
\usepackage{enumitem}
\usepackage{inconsolata}
\usepackage{colortbl}
\usepackage{makecell}
\usepackage[linewidth=1pt]{mdframed}
\usepackage{times}
\usepackage{latexsym}
\usepackage{booktabs}
\usepackage{multirow}
\usepackage[utf8]{inputenc}
\usepackage{pifont}
\usepackage{soul}
\usepackage{comment}
\usepackage[export]{adjustbox}
\usepackage{csquotes}

\usepackage[utf8]{inputenc} 
\usepackage[T1]{fontenc}    
\usepackage{hyperref}       
\usepackage{url}            
\usepackage{booktabs}       
\usepackage{amsfonts}       
\usepackage{nicefrac}       
\usepackage{microtype}      
\usepackage{xcolor}         
\usepackage{standalone}
\usepackage{latexsym}
\usepackage{amsmath}
\usepackage{amssymb}
\usepackage{amsthm}
\usepackage{graphicx}
\usepackage{subcaption}
\usepackage{array}
\usepackage{tabu}
\usepackage{makecell}
\usepackage{paralist}
\usepackage{cases}
\usepackage{diagbox}
\usepackage{enumitem}
\usepackage{soul}
\usepackage{multirow}
\usepackage{verbatim}
\usepackage{tabulary}
\usepackage{booktabs}
\usepackage[mathscr]{euscript}
\usepackage{mathtools}
\usepackage{algorithm}
\usepackage{algpseudocode}
\usepackage{stmaryrd}
\usepackage{tikz-dependency}
\usetikzlibrary{automata,decorations.markings,arrows,positioning,matrix,calc,patterns,angles,quotes,calc}
\usepackage{adjustbox}
\usepackage{tabularx}
\usepackage{xspace}
\usepackage{tabulary}
\usepackage{afterpage}
\usepackage{bm}
\usepackage{color}
\usepackage{graphicx}
\usepackage{slashbox}
\usepackage[toc,page]{appendix}
\usepackage{makecell}
\usepackage{boldline}
\usepackage[shortcuts]{extdash}  

\usepackage{blindtext}
\usepackage{graphicx}
\usepackage{capt-of}
\usepackage{booktabs}
\usepackage{varwidth}
\usepackage{pifont}
\usepackage{wrapfig}

\usepackage{listings}

\usepackage{pythonhighlight}
\usepackage{bm}

\usepackage{graphics}
\usepackage{graphicx}
\usepackage{amsmath}
\usepackage{amssymb}
\usepackage{newfloat}
\usepackage{listings}
\usepackage{diagbox}
\allowdisplaybreaks[4]
\definecolor{orange}{rgb}{1,0.5,0}
\definecolor{mdgreen}{rgb}{0.05,0.6,0.05}
\definecolor{mdblue}{rgb}{0,0,0.7}
\definecolor{dkblue}{rgb}{0,0,0.5}
\definecolor{dkgray}{rgb}{0.3,0.3,0.3}
\definecolor{slate}{rgb}{0.25,0.25,0.4}
\definecolor{gray}{rgb}{0.5,0.5,0.5}
\definecolor{ltgray}{rgb}{0.7,0.7,0.7}
\definecolor{purple}{rgb}{0.7,0,1.0}
\definecolor{lavender}{rgb}{0.65,0.55,1.0}

\definecolor{mypurple}{RGB}{111,61,121}
\definecolor{myblue}{RGB}{46,88,180}
\definecolor{myred}{RGB}{181,68,106}
\definecolor{myyellow}{RGB}{204,143,55}

\title{Language Models Hallucinate, but May Excel at Fact Verification}

\usepackage[multiple]{footmisc}
\usepackage{bigfoot}

\author{
Jian Guan$^{1*}$, Jesse Dodge$^2$, David Wadden$^2$, 
Minlie Huang$^{1\dagger}$, Hao Peng$^{3}$\thanks{~This work was partially done when Jian Guan and Hao Peng were at Allen Institute for AI. \\$~~~~~~~~~\dagger$ Corresponding author.}\\
$^1$The CoAI group, DCST, Institute for Artificial Intelligence, \\
$^1$State Key Lab of Intelligent Technology and Systems,\\
$^1$Beijing National Research Center for Information Science and Technology, \\
$^1$Tsinghua University, Beijing 100084, China. \\
$^2$Allen Institute for AI, $^3$University of Illinois Urbana-Champaign\\
{\texttt{j-guan19@mails.tsinghua.edu.cn}}, 
{\texttt{jessed@allenai.org}}, 
{\texttt{davidw@allenai.org}},\\
{\texttt{aihuang@tsinghua.edu.cn}},
{\texttt{haopeng@illinois.edu}}
 \\
}
\begin{document}
\maketitle
\begin{abstract}
Recent progress in natural language processing~(NLP) owes much to remarkable advances in large language models (LLMs). Nevertheless, LLMs frequently ``hallucinate,’’ resulting in non-factual outputs.
Our carefully-designed human evaluation substantiates the serious hallucination issue, revealing that even GPT-3.5 produces factual outputs less than 25\% of the time. This underscores the importance of fact verifiers in order to measure and incentivize progress. Our systematic investigation affirms that LLMs can be repurposed as effective fact {verifiers} with strong correlations with human judgments. 
Surprisingly, FLAN-T5$_{\textsc{11b}}$, the least factual generator in our study, performs the best as a fact verifier, even outperforming more capable LLMs like GPT3.5 and ChatGPT. Delving deeper, we analyze the reliance of these LLMs on high-quality evidence, as well as their deficiencies in robustness and generalization ability.
Our study presents insights for developing trustworthy generation models.

\end{abstract}

\section{Introduction}
LLMs have demonstrated remarkable performance across 
various natural language generation~(NLG) tasks~\citep{brown2020language,openai2023gpt4,chowdhery2022palm}.
However, they persistently suffer from the \textit{hallucination} problem~\cite{bang2023multitask}, often generating non-factual and sometimes misleading outputs.
This is quantitatively substantiated by the first part of this paper.
In our carefully designed human evaluation of several current LLMs, GPT-3.5 only manages to produce factual outputs less than 25\% of the time;
other models perform even worse.
Such underperformance is achieved on Wikipedia, a domain that they have been extensively trained in and intuitively ``familiar with.''
Our findings highlight the serious challenge that the hallucination issue presents,
and underscore the crucial importance of developing effective fact verification methods~\cite{vlachos-riedel-2014-fact}.
These methods are central to evaluating and incentivizing progress in improving LLMs' factuality. 


In the second part of the paper, we explore the prospect of leveraging instruction-tuned LLMs for fact verification. 
We hypothesize that, despite struggling to generate factual outputs, they may still 
be able to judge whether a piece of text 
is factual
\textemdash a task that intuitively appears easier, at least for sentence-level judgments.
Our systematic investigation affirms this hypothesis, especially when LLMs are augmented with retrieval components. Specifically, given a statement to be verified, we retrieve evidence from an external corpus and re-frame the statement and evidence into a prompt to instruct an LLM to judge the factuality. 
We then normalize the LLM's generation probabilities of pre-defined answers as the factuality score, 
which shows stronger correlations with human judgments than previous statistical and model-based methods. Extensive experiments further reveal that FLAN-T5$_{\textsc{11b}}$~\cite{chung2022scaling}, the least factual generator in our study,  
even surprisingly outperforms GPT3.5 and ChatGPT for fact verification.

We further analyze the LLM-based fact verifiers from the following perspectives: 
\textbf{(1) Influence of given evidence:} 
ChatGPT is susceptible to irrelevant evidence but deals with relevant but counterfactual evidence better than FLAN-T5$_{\textsc{11b}}$. \textbf{(2) Robustness:} GPT variants are less robust to different prompts than FLAN-T5$_{\textsc{11b}}$; \textbf{(3) Generalization ability:} 
It is more difficult to evaluate sentences that are from larger generators, dependent on the context or involving numerals. Evaluating paragraphs is also challenging, and can be facilitated by aggregating judgments of individual, de-contextualized sentences rather than evaluating them directly.
Our contributions are as follows:

\noindent\textbf{I.} Well-designed human evaluation affirms the serious challenges that current LLMs frequently hallucinate, even in their familiar Wikipedia domain. 

\noindent\textbf{II.} We explore the potential of LLMs to assess factuality on multiple domains and analyze their reliance on given evidence, robustness, and generalization ability. These findings may inspire the development of trustworthy generation models and fact verification methods in future research. \footnote{The data and evaluation scripts are publicly available at \url{https://github.com/JianGuanTHU/LLMforFV}.}.
The evaluation suite also serves as a new comprehensive benchmark for hallucination evaluation\footnote{All the data and evaluation scripts will be made public.}.

\noindent\textbf{III.} Based on our study, we recommend the following practices for fact verification: minimizing irrelevant evidence, taking sentences as base units for long paragraph verification, and de-contextualizing context-dependent sentences before verification.

\section{Related Work}
\paragraph{Hallucination} Many metrics have been proposed to measure hallucinations for directed generation tasks such as summarization, including statistical and model-based metrics. Statistical metrics focus on lexical input-output matching~\cite{dhingra2019handling,wang2020towards,shuster2021retrieval}. 
Model-based ones further capture semantic-level variations, including unsupervised metrics based on information extraction~(IE)~\cite{nan2021entity}, question answering~(QA)~\cite{wang-etal-2020-asking} and natural language inference~(NLI)~\cite{laban2022summac}, and supervised or semi-supervised metrics trained on specific datasets of evaluation-related tasks~\cite{izacard2021leveraging,kryscinski2020evaluating}. These metrics can potentially adapt to open-ended generation by measuring mismatching between outputs and retrieved evidence. One additional challenge lies in retrievers possibly producing noisy, redundant, or contradictory evidence. 

\paragraph{Fact Verification} Lots of datasets have been collected towards fact verification in various domains, e.g., politics~\cite{vlachos-riedel-2014-fact}, encyclopedia~\cite{thorne-etal-2018-fever,thorne2021evidence,eisenschlos2021fool}, news~\cite{perez-rosas-etal-2018-automatic}, climate~\cite{diggelmann2020climate}, science~\cite{wadden-etal-2020-fact}, and healthcare~\cite{kotonya-toni-2020-explainable-automated}. \citet{honovich2022true} aggregated multiple datasets to assess the ability to measure input-output consistency. Statements in all above datasets usually contain only single sentences and are crafted by either crawling from 
dedicated websites~\cite{vlachos-riedel-2014-fact}, manually mutating sentences from factual articles~\cite{thorne-etal-2018-fever} or re-framing QA pairs~\cite{thorne2021evidence}. We further involve model-generated statements and paragraph-level evaluation in our study. 

\paragraph{LLMs as Evaluators} There are many active efforts to use LLMs' generated answers or generation probabilities for NLG evaluation~\cite{yuan2021bartscore,colombo2022infolm,ke2022ctrleval}. More recent studies show high correlations of ChatGPT with human judgments for evaluating summarization, story generation, etc.~\cite{wang2023chatgpt,luo2023chatgpt,li2023halueval}. \textsc{SelfCheckGPT}~\cite{manakul2023selfcheckgpt} judged the factuality of a model output based on its similarity with other sampled outputs from the same model, and does not apply to non-model-generated statements or model-agnostic generation.
\textsc{FActScore}~\cite{min2023factscore} used LLMs to evaluate people's biography generation through the fraction of atomic facts supported by retrieved evidence. In contrast, we focus on open-ended generation around various entities and analyze LLMs' robustness and generalization ability.

\section{Quantifying LLMs' Hallucination}\label{quantity}
Our first research question is

\begin{displayquote}
\emph {To what extent do current LLMs hallucinate?}
\end{displayquote}
We quantify this through human evaluation, with a  specific focus on the Wikipedia domain, which serves as a reliable information source for annotators. While clean and credible resources exist for other domains such as science and finance, it remains challenging for individuals lacking the expertise to evaluate LLMs' outputs in these domains.


\begin{table*}[!t]
\small
\centering
\begin{adjustbox}{max width=\linewidth}
\begin{tabular}{@{}p{220pt}m{0.001em}p{360pt}@{}}
\toprule
\textbf{\textsc{SentCom}}&&\textsc{\textbf{ParaGen}}\\
\cline{1-1}
\cline{3-3}
{{Please complete the sentence following the given beginning:}}\newline\newline
{Beginning: Swedish Empire}\newline
{Continuation: was ruled by Gustavus Adolphus from 1611 to 1632 .}\newline
$\cdots$\newline
{Beginning: \textcolor{red}{\{input\}}}\newline
{Continuation:}
&&
{{Please answer the following questions:}}\newline\newline
{Question: Please write five sentences about facts of ``Fire and Darkness''}\newline
{Answer: Fire and Darkness is a cancelled three-dimensional real-time strategy video game developed by Singularity Software. The game consists of a player controlling one of two factions $\cdots$}\newline
$\cdots$\newline
{Question: Please write five sentences about facts of ``\textcolor{red}{\{input\}}''}\newline
{Answer:}\\
\bottomrule
\end{tabular}
\end{adjustbox}
\caption{Prompts for collecting model outputs. \textcolor{red}{{\{Input\}}} is two tokens/an entity for \textsc{SentCom}/\textsc{{ParaGen}}.}
\label{gen_prompt}
\end{table*}

\paragraph{Generation Models} We consider four representative LLMs, 
including FLAN-T5$_{\textsc{11b}}$~\cite{chung2022scaling}, 
LLama$_{\textsc{30b}}$, LLama$_{\textsc{65b}}$~\cite{touvron2023llama} and GPT3.5\footnote{In this work, InstructGPT/GPT3.5/ChatGPT/ refers to the OpenAI's API ``text-davinci-002''/``text-davinci-003''/``gpt-3.5-turbo-0301,'' respectively.}. These LLMs vary in model architectures, model sizes, accessibility and training manners. We expect to establish a clear relationship between these variables and final hallucination performance.

\paragraph{Generation Tasks} We design two open-ended generation tasks that simulate realistic interactions between practitioners and LLMs: 
\textbf{(1) Sentence Completion~{(\textsc{SentCom})}:} 
completing a sentence following the first two tokens of a factual claim from the test set of FEVER~\cite{thorne-etal-2018-fever}, a fact verification dataset. 
The training set provides abundant factual and non-factual claims, 
enabling us to assess whether supervised verifiers can generalize to model outputs in \S\ref{repurpose}. 
\textbf{(2) Wikipedia Paragraph 
Generation~(\textsc{ParaGen})}: generating a paragraph of five sentences about a given entity from Wikipedia.
Here the outputs are expected to be longer, and the setting focuses on more long-tailed topics than (1). 

We generate 50 outputs for two tasks, respectively, using four generation models with greedy decoding\footnote{\citet{aksitov2023characterizing} showed that lower temperatures generally lead to less variability and potentially higher factuality.}, leading to $50\times2\times4=400$ 
statements totally. During generation, we provide five manually selected factual demonstrations to the models, as shown in Tab.~\ref{gen_prompt}.

\begin{figure}[!t]
  \centering
\includegraphics[width=\linewidth]{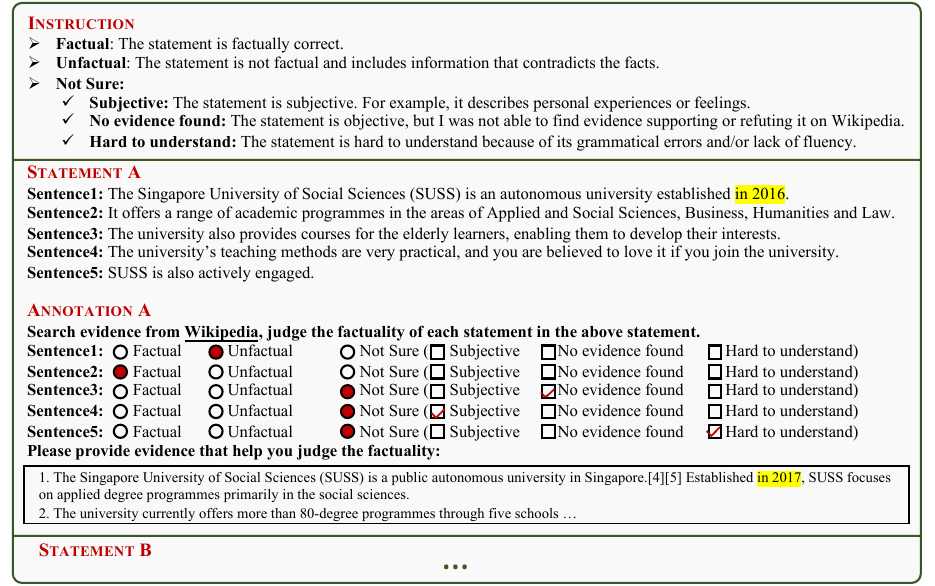}
  \caption{The annotation interface for \textsc{ParaGen}; that for \textsc{SentCom} is similar.} 
  \label{fig:ant}
\end{figure}

\begin{table}[!t]
\small
\centering
\begin{adjustbox}{max width=\columnwidth}
\begin{tabular}{@{}ll rrrr@{}}
\toprule
\multirow{2}{*}{\textbf{Outputs}}&\multirow{2}{*}{\textbf{Models}}&\multirow{2}{*}{\textbf{\#}}&\multicolumn{3}{c}{\textbf{Proportion~(\%)}}\\
\cline{4-6}
&&&\textbf{\texttt{Factual}}&\textbf{\texttt{Unfactual}}&\textbf{\texttt{NE}}\\
\midrule
\multirow{4}{*}{{\textbf{\textsc{SentCom}}}}&\textbf{FLAN-T5$_{\textsc{11b}}$}&48&33.3&50.0&16.7\\
&\textbf{Llama$_{\textsc{30b}}$}&49&\underline{75.5}&14.3&10.2\\
&\textbf{Llama$_{\textsc{65b}}$}&47&68.1&19.2&12.8\\
&\textbf{GPT3.5$_{\textsc{175b}}$}&49&\textbf{89.8}&6.1&4.1\\
\midrule
&\textbf{FLAN-T5$_{\textsc{11b}}$}&147&10.2&58.5&31.3\\
{{\textbf{\textsc{ParaGen}}}}&\textbf{LLama$_{\textsc{30b}}$}&143&29.4&41.3&29.4\\
{{\textbf{(Sent)}}}&\textbf{LLama$_{\textsc{65b}}$}&139&\underline{33.1}&36.0&30.9\\
&\textbf{GPT3.5$_{\textsc{175b}}$}&139&\textbf{37.4}&37.4&25.2\\
\midrule
&\textbf{FLAN-T5$_{\textsc{11b}}$}&25&0.0&92.0&8.0\\
{{\textbf{\textsc{ParaGen}}}}&\textbf{Llama$_{\textsc{30b}}$}&25&4.0&80.0&16.0\\
{{\textbf{(Para)}}}&\textbf{Llama$_{\textsc{65b}}$}&21&\underline{9.5}&85.7&4.8\\
&\textbf{GPT3.5$_{\textsc{175b}}$}&22&\textbf{22.7}&68.2&9.1\\
\bottomrule
\end{tabular}
\end{adjustbox}
\caption{Statistics of model outputs. 
\textbf{Sent/Para:} {Sentence/Paragraph}; \textbf{\texttt{NE:}} \texttt{{No Evidence}}. {\bf \#}: the number of annotated instances. \textbf{Bold}/\underline{Underlined} percentages indicate the \textbf{most}/\underline{second most} factual outputs.} 
\label{gen_stat}
\end{table}

\paragraph{Human Annotation} 
We collect human workers' factuality judgments of LLMs' outputs through Amazon Mechanical Turk~(AMT). 
Each HIT~(human intelligence task) contains five statements with the same input\textemdash four generated and one gold. 
Three workers are hired to search for evidence from Wikipedia and annotate the factuality label of each sentence in the statements, including \texttt{factual}, \texttt{unfactual} and \texttt{not sure}, as shown in Fig.~\ref{fig:ant}\footnote{We only take Wikipedia as the reliable information source since 
there is much noisy, biased, and non-validated information on the Internet~\cite{shu2017fake}.}. 
We further instruct workers to choose reasons if they annotate \texttt{not sure}, and exclude sentences from our evaluation set that are labeled as 
\texttt{subjective} or \texttt{hard to understand} by at least one worker since such sentences can be 
ambiguous to determine the factuality~\cite{guo2022survey}. 
We discard low-quality submissions using well-designed rules and ensure each sentence has three valid annotations. The Fleiss's kappa score~\cite{Fleiss1971Measuring} is 0.91/0.74 for \textsc{SentCom}/\textsc{ParaGen}, indicating a substantial 
inter-annotator agreement.
Finally, we use majority voting to obtain sentence-level human judgments. As for paragraph-level judgments for \textsc{ParaGen}, we first truncate each paragraph to ensure all sentences are not subjective or hard to understand, 
and then label the paragraph \texttt{unfactual} if \emph{any} of the sentences are labeled \texttt{unfactual}, \texttt{factual} if all sentences are labeled \texttt{factual}, and \texttt{not sure} otherwise. A paragraph contains 4.3 sentences on average after truncation.
Since the only valid reason for  \texttt{not sure} is  \texttt{no evidence found}, we call the label  \texttt{no evidence} onward. 
Appendix~\ref{man_ant} presents more details. 

\begin{figure}[!t]
  \centering
\includegraphics[width=\linewidth]{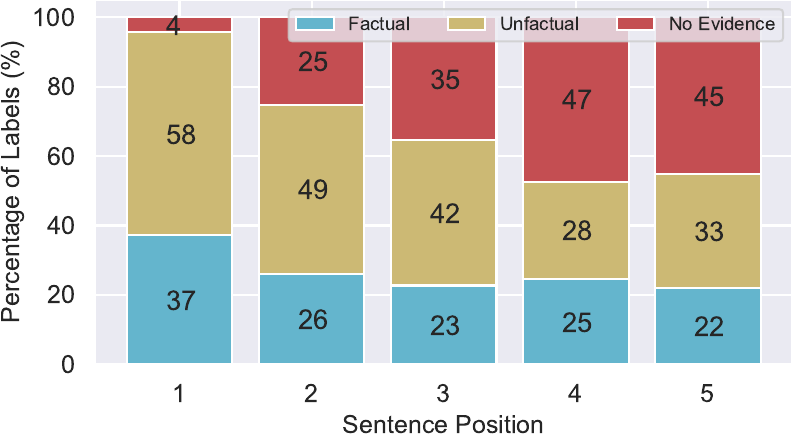}
  \caption{Distribution of different labels arcross sentence positions in \textsc{ParaGen}~(Para).} 
  \label{fig:prop}
\end{figure}

\begin{table*}[!t]
\centering
\begin{adjustbox}{max width=\textwidth}
\begin{tabular}{@{} l ccc m{0.001em} cccc @{}}
\toprule
\multirow{2}{*}{\textbf{Datasets}}&\multicolumn{3}{c}{\textbf{\textsc{wks}}}&&\multicolumn{4}{c}{\textbf{\textsc{dss}}}\\
\cline{2-4}
\cline{6-9}
&\textbf{FEVER}&\textbf{BoolQ-FV}&\textbf{FM2}&&\textbf{PubMedQA}&\textbf{XsumFaith}&\textbf{SummEval}&\textbf{SciFact}\\
\midrule
\textbf{\# Examples}&1,000&613&1,380&&445&853&798&191\\
\textbf{\# Factual}&517&433&681&&276&60&719&101\\
\textbf{\# Unfactal}&483&180&699&&169&793&79&90\\
\textbf{Avg. Len}&9.38&9.57&15.34&&19.07&25.1&76.53&12.80\\
\midrule
\multirow{2}{*}{\textbf{Source}}
&Wikipedia&Search-engine&Adversarial&&\multirow{2}{*}{PubMed}&\multirow{2}{*}{BBC}&\multirow{2}{*}{CNN/DailyMail}&Scientific\\
&Articles&Queries&Games&&&&&Papers\\
\midrule
\textbf{Domain}&Wikipedia&Wikipedia&Wikipedia&&Medicine&News&News&Science\\
\bottomrule
\end{tabular}
\end{adjustbox}
\caption{Statistics of test sets of three datasets in \textsc{wks} and four datasets in \textsc{dss}.} 
\label{data_stat}
\end{table*}

Tab.~\ref{gen_stat} summarizes the annotation results: 
\textbf{(1)} Larger models tend to generate more factual outputs. \textbf{(2)} 
GPT3.5, the best-performing generator in this study, yields factual paragraphs less than 25\% of the time. 
\textbf{(3)}
All models generate notably more factual outputs for \textsc{SentCom} than \textsc{ParaGen}. 
We conjecture it is because the average frequency of inputs for \textsc{SentCom} is $\sim$335 times more than that of input entities for \textsc{ParaGen}. We count frequency using the WebText corpus~\cite{radford2019language}. \textbf{(4)} Fig.~\ref{fig:prop} shows increasing percentages of \texttt{No Evidence} 
as the generation proceeds to later sentences, 
which may be caused by irrelevant or spurious information introduced into outputs by the error accumulation 
inherent in auto-regressive generation~\cite{zhang2023language}. Appendix~\ref{context_fact} shows the influence of the context on model generation and fact verification. 

\textcolor{black}{The above human evaluation confirms the LLMs' serious hallucination issue, emphasizing the urgent need for effective fact verifiers to measure and incentivize progress in LLMs' factuality. This motivates us to explore LLMs' potential as fact verifiers.}

\section{Repurposing LLMs as Fact Verifiers}\label{repurpose}
Our second inquiry lies around the question
\begin{displayquote}
    \emph{Can LLMs be repurposed as effective fact verifiers?}
\end{displayquote}
We define fact verification as follows: given a statement $s$ and its leading context $c$, the verifier should give a probability $p$  
of $s$ being factual\footnote{We formulate the fact verification task as regression instead of traditional classification because continuous scores are more informative to reflect the nuance of inputs and may give generators fine-grained feedback in future work.}. 
The context may be absent, and the statement is a sentence or paragraph. 
Additionally, we retrieve an evidence set, denoted as $E=\{e_1,e_2,\cdots,e_M\}$, from external corpora~\cite{lewis2020retrieval} 
with the concatenation 
of $c$ and $s$ as the query. 
Each piece of evidence is a passage. 
Next, we first describe the evaluation sets~(\S\ref{evaluation_set}), our verification method~(\S\ref{verification_method}), compared verifiers~(\S\ref{implementation}), and then present the experiment results~(\S\ref{results}). 

\begin{table}[!t]
\small
\centering
\begin{adjustbox}{max width=\columnwidth}
\begin{tabular}{@{}lp{222pt}@{}}
\toprule
\textcolor{red}{(Task)}&{{Answer the following question:}}\\
\textcolor{red}{(Input)}&{Facts:\newline1. \{$e_1$\}\newline2. \{$e_2$\}\newline$\cdots$\newline M. \{$e_M$\}}\newline
{Context: \{$c$\}}\newline
{Statement following the context: \{$s$\}}\\
\textcolor{red}{(Question)}&{{Based on the given facts, is the statement correct? (A) Yes. (B) No. Please answer A or B:}}\\
\bottomrule
\end{tabular}
\end{adjustbox}
\caption{An example prompt used to adapt LLMs for fact verification. The prompt may be changed under different settings. For example, when $c$ is empty, we delete  ``{Context}: \{{\textit{c}}\}'' and ``{following the context}''.}
\label{ipt_prompt}
\end{table}

\begin{table*}[!ht]
\centering
\begin{adjustbox}{max width=0.85\textwidth}
\begin{tabular}{@{} l cccc m{0.001em} cccc m{0.001em} cccc@{}}
\toprule
\multirow{2}{*}{\textbf{Verifiers}}
& \multicolumn{4}{c}{\textbf{FEVER}} 
&& \multicolumn{4}{c}{\textbf{BoolQ-FV}} 
&& \multicolumn{4}{c}{\textbf{FM2}} \\
\cline{2-5}
\cline{7-10}
\cline{12-15}
&\textbf{ECE}
&\textbf{ACC}
&\textbf{AUR}
&\textbf{$r$}
&
&\textbf{ECE}
&\textbf{ACC}
&\textbf{AUR}
&\textbf{$r$}
&
&\textbf{ECE}
&\textbf{ACC}
&\textbf{AUR}
&\textbf{$r$}\\
\midrule
\textbf{Constant} 
& \textit{48.3}
& \textit{51.7}
& \textit{50.0}
& \textit{N/A}
&
& \textit{29.4}
& \textit{70.6}
&\textit{ 50.0}
& \textit{N/A}
&
&\textit{ 50.6}
& \textit{49.4}
& \textit{50.0}
& \textit{N/A}
\\
\midrule
\multicolumn{15}{c}{{\cellcolor[gray]{.95}}\textbf{Retrieving Evidence from External Corpora}}\\
\hline
\textbf{KF1}
& 51.3
& 48.3
& 53.9
& 9.2
&
& 70.3
& 29.4
& 44.4
& {-7.7}
&
& 48.8
& 50.6
& 49.8
& {-0.6}\\
\textbf{NLI$_{\textsc{11b}}$}
& 18.3
& 81.7
& 83.8
& 67.4
&
& 45.1
& 54.8
& 66.8
& 33.2
&
& 34.5
& 65.5
& 68.0
& 38.7\\

\textbf{FiD$_{\textsc{780m}}$}
& \textbf{2.9}
& \textbf{94.6}
& \textbf{98.2}
& \textbf{90.5}
&
& 13.8
& 82.7
& \underline{88.5}
& 62.0
&
& \underline{15.5}
& 77.0
& \underline{85.6}
& \underline{59.1}\\

\hline
\textbf{FLAN-T5$_{\textsc{11b}}$}
& \underline{3.1}
& \underline{93.8}
& \underline{98.2}
& \underline{90.2}
&
& \textbf{10.2}
& \underline{85.5}
& \textbf{94.7}
& \textbf{75.3}
&
& \textbf{8.2}
& \textbf{82.0}
& \textbf{89.5}
& \textbf{68.8}
\\

\textbf{GPT3.5}
& 7.6
& 91.7
& 96.6
& 84.7
&
& 17.9
& 81.7
& 87.4
& 61.1
&
& 21.0
& 77.5
& 82.8
& 55.9
\\
\textbf{ChatGPT}
& \textcolor{black}{8.2}
& 92.8
& \textcolor{black}{92.6}
& \textcolor{black}{84.3}
&
& \textcolor{black}{\underline{13.1}}
& \textbf{87.3}
& \textcolor{black}{88.0}
& \textcolor{black}{\underline{70.1}}
&
& \textcolor{black}{21.4}
& \underline{79.1}
& \textcolor{black}{78.6}
& \textcolor{black}{56.7}
\\
\midrule
\multicolumn{15}{c}{{\cellcolor[gray]{.95}}\textbf{Not Using any Evidence}}\\
\hline
\textbf{FiD$_{\textsc{780m}}$}
& \textbf{4.2}
& {77.3}
& \underline{85.7}
& {62.0}
&
& \textbf{17.4}
& {64.6}
& 60.8
& 17.2
&
& \underline{21.6}
& {59.4}
& {65.1}
& {26.4}
\\
\hline

\textbf{FLAN-T5$_{\textsc{11b}}$}
& \underline{11.1}
& \underline{74.4}
& \textbf{87.0}
& \underline{62.6}
&
& 28.6
& 56.0
& {65.6}
& \underline{23.6}
&
& \textbf{15.7}
& \underline{59.5}
& \underline{66.3}
& \underline{28.2}
\\

\textbf{GPT3.5}
& 25.6
& 73.8
& 78.3
& 52.7
&
& 32.8
& \underline{65.6}
& \underline{67.9}
& {23.2}
&
& 41.4
& 57.3
& 61.5
& 19.5
\\
\textbf{ChatGPT}
& \textcolor{black}{18.1}
& \textbf{82.0}
& \textcolor{black}{81.5}
& \textcolor{black}{\textbf{63.9}}
&
& \textcolor{black}{\underline{27.1}}
& \textbf{73.1}
& \textcolor{black}{\textbf{69.7}}
& \textcolor{black}{\textbf{37.4}}
&
& \textcolor{black}{33.5}
& \textbf{66.7}
& \textcolor{black}{\textbf{66.5}}
& \textcolor{black}{\textbf{34.1}}
\\


\bottomrule
\end{tabular}
\end{adjustbox}
\caption{
Results on \textsc{wks}. ``Constant'' always predicts a factuality score of 1. 
All instruction-tuned LLMs are under the zero-shot setting. 
We highlight the best result in \textbf{bold} and \underline{underline} the second best. Note that the results of ChatGPT except for ACC may be underestimated 
because they are calculated based on generated answers instead of probabilities~(details in Appendix~\ref{chatgpt}). \textbf{Takeaway:} In the Wikipedia domain, ChatGPT performs the best when not using any evidence, while FALN-T5$_{\textsc{11b}}$ excels ChatGPT with retrieved evidence.}
\label{human_eval_stat}
\end{table*}

\begin{table*}[!ht]
\centering
\begin{adjustbox}{max width=\textwidth}
\begin{tabular}{@{} l cccc m{0.001em} cccc m{0.001em} cccc m{0.001em} cccc@{} }
\toprule
\multirow{2}{*}{\textbf{Verifiers}}
& \multicolumn{4}{c}{\textbf{PubMedQA}} 
&& \multicolumn{4}{c}{\textbf{XsumFaith}} 
&& \multicolumn{4}{c}{\textbf{SummEval}} 
&& \multicolumn{4}{c}{\textbf{SciFact}}\\
\cline{2-5}
\cline{7-10}
\cline{12-15}
\cline{17-20}
&\textbf{ECE}
&\textbf{ACC}
&\textbf{AUR}
&\textbf{$r$}
&
&\textbf{ECE}
&\textbf{ACC}
&\textbf{AUR}
&\textbf{$r$}
&
&\textbf{ECE}
&\textbf{ACC}
&\textbf{AUR}
&\textbf{$r$}
&
&\textbf{ECE}
&\textbf{ACC}
&\textbf{AUR}
&\textbf{$r$}\\
\midrule
\textbf{FLAN-T5$_{\textsc{11b}}$}&\textbf{12.6}&78.4&\textbf{84.6}&\textbf{60.0}&&\textbf{20.1}&\textbf{78.2}&\textbf{70.8}&20.0&&\textbf{3.5}&92.4&\textbf{93.2}&\textbf{62.0}&& \textbf{11.1}& {83.2}& \textbf{95.3}& \textbf{77.9}\\
\textbf{ChatGPT}&20.4&\textbf{79.6}&76.7&55.7&&23.6&76.4&68.1&\textbf{21.4}&&7.5&\textbf{92.5}&71.0&51.4&&16.5&\textbf{88.0}& 86.4&69.5\\
\bottomrule
\end{tabular}
\end{adjustbox}
\caption{Results on \textsc{dss}. Both models are under the zero-shot setting and provided with the golden evidence. \textbf{Takeaway:} FLAN-T5$_{\textsc{11b}}$ surpasses ChatGPT in terms of most metrics on specific domains.}
\label{human_eval_stat_other}
\end{table*}

\subsection{Evaluation Sets}\label{evaluation_set}
We design three evaluation sets across multiple domains and sources:
\begin{compactitem}
\item[\textbf{(1)}] \textbf{{Model-Generated Statements~(\textsc{mgs})}:} It includes \textsc{SentCom} and \textsc{ParaGen} statements generated by four LLMs in \S\ref{quantity}.\looseness=-1

\item[\textbf{(2)}] \textbf{{Wiki-Domain Statements~(\textsc{wks})}:} It aggregates three fact verification datasets in the Wikipedia domain including FEVER~\cite{petroni2021kilt}, BoolQ-FV~\cite{thorne2021evidence} and FM2~\cite{eisenschlos2021fool}.
\item[\textbf{(3)}] \textbf{Domain-Specific Statements~(\textsc{dss})}: It aggregates four domain-specific datasets including PubMedQA~\cite{jin2019pubmedqa}, XsumFaith~\cite{maynez2020faithfulness}, SummEval~\cite{fabbri2021summeval} and SciFact~\cite{wadden-etal-2020-fact}.
Statements supported/refuted by golden evidence are labeled \texttt{factual}/\texttt{unfactual}, 
and we remove statements without golden evidence since the factuality is unknowable. 

\end{compactitem}
These evaluation sets span various domains, origination, and lengths. Tab.~\ref{data_stat} summarizes the statistics of \textsc{wks} and \textsc{dss}, and Appendix~\ref{collect_hcs} includes more details.  




\subsection{Verification Method}\label{verification_method}
We transform each input $x=(E,c,s)$ into a prompt, 
as shown in Tab.~\ref{ipt_prompt}. 
The LLM is expected to generate a judgment $W=(w_1,w_2,\cdots,w_T)$ about whether $s$ is factual. We compute the factuality score $p$ by normalizing the LLM's output probabilities of all valid answers~\cite{ke2022ctrleval}:
\begin{align*}   \label{metric_score}
p=\frac{\sum\nolimits_{w\in L_A}p_{\textsc{llm}}\left(w \mid x, W_{<t}\right)}{\sum\nolimits_{w\in L_A\cup L_B}p_{\textsc{llm}}\left(w \mid x, W_{<t}\right)},
\end{align*}
where $p_{\textsc{llm}}$ is the LLM's probability distribution over the vocabulary, $L_A$/$L_B$ is the set of plausible answer words to indicate factual/non-factual, and $t$ is the maximum time step that makes each token in $W_{<t}$ not included in $L_A$ or $L_B$. 
We define $L_A=\{$``A'', ``a'', ``Yes'', ``yes'', ``YES''$\}$ and $L_B=\{$``B'', ``b'', ``No'', ``no'', ``NO''$\}$. If $W$ does not include any valid answer words, we set $p$ to 0.5. In Appendix~\ref{verification_method2}, we compare other verification methods such as Chain-of-Thought prompting~\cite{weichain} and Likert-scale rating.

\subsection{Compared Verifiers}\label{implementation}
We test the following instruction-tuned LLMs for fact verification using the method in \S\ref{verification_method}:
\begin{compactitem}
    \item[\textbf{(1)}]
    \textbf{FLAN-T5$_{\textsc{11b}}$:} It is instruction-tuned from T5$_{\textsc{11b}}$~\cite{raffel2020exploring} on 1.8K+ tasks. 
    \item[\textbf{(2)}]
    \textbf{GPT3.5:} We approximate $p_{\textsc{llm}}$ by normalizing the probabilities calculated from top five logits 
    returned by the API. 
    \item[\textbf{(3)}]
    \textbf{ChatGPT}: We hard code the score as 1/0 if it returns ``A''/``B'' using the prompt in Tab.~\ref{ipt_prompt} since its output probabilities are unavailable.  
\end{compactitem}
For all models, the maximum length is set to 4,000 tokens tokenized by the GPT-series BPE tokenizer. Appendix~\ref{more_llms} presents the results of more LLMs. 

We compare the LLMs to baselines widely used to measure hallucination: 
\begin{compactitem}
    \item[\textbf{(1)}]
    \textbf{knowledgeF1~(KF1)}: It 
    measures the average unigram overlap between the statement and each piece of evidence~\cite{shuster2021retrieval}. 
    \item[\textbf{(2)}]
    \textbf{NLI}: It is an unsupervised verifier computed as the entailment probability between the evidence and statement. We use the public T5$_{\rm\textsc{11b}}$-based NLI model fine-tuned on a mixture of multiple NLI datasets~\cite{honovich2022true}.  
    \item[\textbf{(3)}]
    \textbf{FiD}: It is a supervised verifier based on a fine-tuned binary factuality classifier~\cite{izacard-grave-2021-leveraging,liu2022token} 
    with the statement and evidence as input. 
    We build FiD based on FLAN-T5$_{\rm \textsc{780m}}$ and fine-tune it on three \textsc{wks} training sets, respectively, to obtain corresponding verifiers.  
    \item[\textbf{(4)}]
    \textbf{\textsc{\textsc{FActScore}}~(FAS)}: It first automatically splits a statement into multiple atomic facts 
    and then computes the factual precision as the overall score, i.e., the percentage of atomic facts supported by evidence~\cite{min2023factscore}.  

\end{compactitem}

Regarding the retrieval components, 
we employ the Wikipedia dump from~\citet{izacard2022unsupervised} as the external corpus for \textsc{mgs} and \textsc{wks}, 
with each piece of evidence being a passage of 100 words. Ten pieces of evidence are retrieved for each test sample using Contriever~\cite{izacard-grave-2021-leveraging}. Appendix~\ref{retriever} presents details about retrievers. For \textsc{dss}, we use the golden evidence provided by the original papers.

\subsection{Results}\label{results}
\paragraph{Results on \textsc{wks}\&\textsc{dss}}
Taking human judgments of factual/unfactual statements as 1/0, we use the following metrics to evaluate fact verifiers (Appendix~\ref{evaluation_metrics} shows more details):
\begin{compactitem}
    \item[\textbf{(1)}]
    \textbf{Expected Calibration Error~(ECE)}~\cite{guo2017calibration}: It estimates to what extent the predicted score can indicate accuracy. Lower ECE mean better calibration. 
    \item[\textbf{(2)}]
    \textbf{\textbf{Accuracy~(ACC)}}: It is the fraction of examples that are correctly predicted. 
    \item[\textbf{(3)}]
    \textbf{Area Under the ROC Curve~(AUR)}: It measures the ability to discriminate factual statements from others. 
    \item[\textbf{(4)}]
    \textbf{Pearson's Correlation~($r$)}: It measures the correlation between prediction and human judgments. 
\end{compactitem}


As shown in Tab.~\ref{human_eval_stat}, on \textsc{wks}, \textbf{when using retrieved evidence}, \textbf{(1)} 
KF1 hardly captures any factuality features; \textbf{(2)} NLI$_{\textsc{11b}}$ struggles to generalize to fact verification and significantly underperforms FLAN-T5$_{\textsc{11b}}$; \textbf{(3)} Instruction-tuned LLMs achieve strong performance 
and are comparable to or better than supervised FiD models; 
And \textbf{(4)} FLAN-T5$_{\textsc{11b}}$ outperforms GPT3.5 in both calibration and discrimination ability. 
\textbf{When not using any evidence}, 
ChatGPT performs the best, possibly attributed to its superiority in memorizing and utilizing knowledge. 
The overall inferior performance of not using evidence reveals the importance of retrieval components. Furthermore, the results on \textsc{dss} in Tab. \ref{human_eval_stat_other} show that FLAN-T5$_{\textsc{11b}}$ surpasses ChatGPT in terms of most metrics, which also indicates the potential of relatively small-scaled models for hallucination evaluation across various domains. Appendix~\ref{factprompts} shows a similar conclusion on a contemporary dataset FactPrompts~\cite{chern2023factool}.\looseness=-1

\paragraph{Results on \textsc{mgs}}
Besides metrics used on \textsc{wks}, we also report \textbf{Precision~(P)}, \textbf{Recall~(R)}, and \textbf{Area Under the Precision-Recall Curve~(AUP)} on the \texttt{factual} category for \textsc{SentCom} and \textsc{ParaGen}~(Sent), considering the label imbalance. 
We treat human judgments of factual sentences as 1 and others as 0\footnote{Since neither ``No Evidence'' nor ``Unfactual'' is tolerable for users, we do not distinguish the two categories.}, and calculate the human judgment of a paragraph as the faction of factual sentences.
We use FiD trained on FEVER for the experiments.

\begin{table}[!t]
\centering
\begin{adjustbox}{max width=\columnwidth}
\begin{tabular}{@{} l cccccc @{}}
\toprule
\textbf{Verifiers}&\textbf{ECE}&\textbf{P}&\textbf{R}&\textbf{AUR}&\textbf{AUP}&\textbf{$r$}\\
\midrule
\textbf{Constant} & \textit{33.2} & \textit{66.8} & \textit{100.0} & \textit{50.0} & \textit{66.8} & \textit{N/A} \\
\textbf{FiD} & \underline{11.0} & 79.0 & 96.1 & 88.6 & \underline{93.7} & 64.2 \\
\hline
\textbf{FLAN-T5$_{\rm\textsc{11b}}$} & \textbf{10.5} & \textbf{96.5} & 84.5 & \textbf{93.1 }& \textbf{96.9} & \textbf{74.9} \\
\textbf{GPT3.5} & 17.7 & 85.2 & \underline{89.2} & 87.8 & 90.8 & 60.8 \\
\textbf{ChatGPT} & \textcolor{black}{15.0} & \underline{91.7} & 85.3 & \textcolor{black}{84.8} & \textcolor{black}{88.0} & \textcolor{black}{\underline{67.6}} \\
\hline
\textbf{{FAS}$_{\textsc{FLAN-T5}}$} & 11.7 & \underline{91.7} & 77.5 & \underline{89.8} & \underline{93.7} & 69.3 \\
\textbf{{FAS}$_{\textsc{chat}}$} & 18.5 & 80.4 & \textbf{92.2} & 78.8 & 83.6 & 55.0 \\

\bottomrule
\end{tabular}
\end{adjustbox}
\begin{adjustbox}{max width=\columnwidth}
\begin{tabular}{@{} l cccccc m{0.001em} c @{} }
\toprule
\multirow{2}{*}{\textbf{Verifiers}}
&\multicolumn{6}{c}{\textbf{Sent}}
&&\multicolumn{1}{c}{\textbf{Para}}\\
\cline{2-7}
\cline{9-9}
&\textbf{ECE}
&\textbf{P}
&\textbf{R}
&\textbf{AUR}
&\textbf{AUP}
&\textbf{$r$}
&&\textbf{$r$}\\
\midrule
\textbf{Constant}
& \textit{72.7}
& \textit{27.3}
& \textit{100.0}
& \textit{50.0}
& \textit{27.3}
& \textit{N/A}
& & \textit{N/A}\\
\textbf{FiD}
& 38.9&35.9&\underline{92.9}&85.9&\underline{76.7}&46.5
& & 39.1\\
\hline
\textbf{FLAN-T5$_{\rm\textsc{11b}}$}
& \textbf{9.2}
& \underline{76.0}
& 71.6
& \textbf{88.0}
& \textbf{80.9}
& \textbf{66.8}
&&45.1 / \underline{79.4}\\
\textbf{GPT3.5}
& 41.4
& 38.5
& \textbf{95.5}
& \underline{87.7}
& 70.6
& 39.5
&& 48.6 / 59.7\\
\textbf{ChatGPT}
& \textcolor{black}{25.6}
& 51.7
& {90.3}
& \textcolor{black}{79.7}
& \textcolor{black}{49.6}
& \textcolor{black}{52.6}
&& \textcolor{black}{52.6} / \textcolor{black}{68.1}\\
\hline
\textbf{FAS$_{\textsc{FLAN-T5}}$}
& \underline{10.6}
& \textbf{77.4}
& 57.4
& 84.0
& {72.6}
& \underline{61.9}
&& \textbf{79.7}\\
\textbf{{FAS}$_{\textsc{chat}}$}
& \textcolor{black}{45.5}
& {35.7}
& {85.2}
& \textcolor{black}{70.7}
& \textcolor{black}{42.4}
& \textcolor{black}{29.4}
&&45.3\\
\bottomrule
\end{tabular}
\end{adjustbox}
\caption{Results on  \textsc{mgs} (\textbf{Top:} \textsc{SentCom}; \textbf{Bottom:} \textsc{ParaGen}) with retrieved evidence. 
Two scores \textbf{A}/\textbf{B} of LLMs on \textsc{ParaGen}~(Para) mean, \textbf{A}: directly evaluating a whole paragraph, \textbf{B}: averaging corresponding sentence-level scores. FAS$_{\textsc{FLAN-T5/chat}}$ refers to \textsc{FAS} using FLAN-T5$_{\textsc{11b}}$/ChatGPT to judge the factuality of atomic facts. \textbf{Takeaway:} FLAN-T5$_\textsc{11b}$ exhibits the overall best performance; Taking sentences as base units for verification, as opposed to paragraphs or potentially noisy atomic facts, yields better results.}
\label{mgs_stat}
\end{table}

\begin{table}[!t]
\small
\centering
\begin{adjustbox}{max width=\columnwidth}
\begin{tabular}{@{} lccccc @{}}
\toprule
\multirow{2}{*}{\textbf{Verifiers}}&\multirow{2}{*}{\textbf{FEVER}}&\multirow{2}{*}{\textbf{FM2}}&\multirow{2}{*}{\textbf{\textsc{SentCom}}}&\multicolumn{2}{c}{\textbf{\textsc{ParaGen}}}\\
\cline{5-6}
&&&&\textbf{Sent}&\textbf{Para}\\
\midrule
\textbf{FLAN-T5$_\textsc{780m}$}&85.8&61.6&\underline{75.4}&55.6&71.8\\
\textbf{FLAN-T5$_\textsc{3b}$}&\underline{88.6}&\underline{66.6}&\textbf{75.8}&\underline{65.8}&\underline{78.6}\\
\textbf{FLAN-T5$_\textsc{11b}$}&\textbf{90.2}&\textbf{68.8}&74.9&\textbf{66.8}&\textbf{79.4}\\
\bottomrule
\end{tabular}
\end{adjustbox}
\caption{Pearson's correlation scores of FLAN-T5 with different model sizes. We calculate factuality scores on \textsc{ParaGen}~(Para) by averaging sentence-level scores here and below.}
\label{model_size}
\end{table}

Tab.~\ref{mgs_stat} shows: \textbf{(1)} Despite FiD's best performance on FEVER, it underperforms 
FLAN-T5$_{\rm\textsc{11b}}$ on \textsc{SentCom}, indicating limited generalization of supervised models. 
\textbf{(2)} All verifiers exhibit inferior performance on \textsc{ParaGen}~(Sent) compared to \textsc{SentCom}, potentially due to less prevalent entities and more contextual dependencies~(e.g., coreference). \textbf{(3)} 
The high recall scores of GPT variants in contrast to FLAN-T5$_{\textsc{11b}}$ show that they prefer to predict \texttt{factual}. This may account for the greater disparity in precision and AUP between GPT variants and FLAN-T5$_{\textsc{11b}}$ on \textsc{ParaGen} that contains more non-factual statements than \textsc{SentCom}. \textbf{(4)} On \textsc{ParaGen}~(Para), averaging sentence-level scores usually yields better correlations than direct verification. 
This is attributed to a stronger ability to capture subtle details within a paragraph, facilitated by independent retrieval and verification for every sentence. 
\textbf{(5)} \textsc{FAS} shows comparable or inferior performance in contrast to FLAN-T5$_{\textsc{11b}}$, suggesting that noise introduced during the generation of atomic facts may impact the final performance.\looseness=-1

\begin{table}[!t]
\centering
\begin{adjustbox}{max width=\columnwidth}
\begin{tabular}{@{} lcc m{0.001em} cc @{}}
\toprule
\multirow{2}{*}{\textbf{Evidence}} & \multicolumn{2}{c}{\textbf{FEVER}} &&\multicolumn{2}{c}{\textbf{FM2}}\\
\cline{2-3}
\cline{5-6}
&\textbf{F-T5}&\textbf{ChatGPT}&&\textbf{F-T5}&\textbf{ChatGPT}\\
\midrule
\textbf{None (0)}&74.4&\textbf{82.0}&&59.5&\textbf{66.7}\\
\textbf{Golden (1)}&93.3&\textbf{94.4}&&87.3&\textbf{88.5}\\
\textbf{Random (10)}&\textbf{64.8}&61.5&&\textbf{56.7}&55.9\\
\textbf{Random+Golden (10)}&\textbf{92.6}&91.5&&\textbf{85.8}&83.5\\
\hline
\textbf{BM25 (10)}&\textbf{87.5}&87.4&&69.1&\textbf{69.3}\\
\textbf{Contriever (10)}&\textbf{93.8}&92.8&&\textbf{82.0}&79.1\\
\hline
\textbf{Adv (1)}&9.4&\textbf{24.0}&&3.4&\textbf{28.2}\\
\textbf{Adv+Golden (2)}&48.1&\textbf{58.9}&&29.9&\textbf{43.7}\\
\bottomrule
\end{tabular}
\end{adjustbox}
\caption{Accuracy of FLAN-T5$_{\textsc{11b}}$~(F-T5) and ChatGPT with different evidence. The numbers in the parentheses are the total number of evidence passages. \textbf{None/Golden:} null/golden evidence; \textbf{Random:} randomly sampled ten passages from Wikipedia; \textbf{Adv}: 
a sentence adversarial with the statement to be verified, constructed by prompting ChatGPT to convert the statement to its antonym/synonym if the statement is factual/unfactual.
}
\label{retriever_analysis}
\end{table}

\paragraph{Influence of Model Sizes}\label{model_size_sec}
Tab.~\ref{model_size} demonstrates a positive correlation between model sizes and the performance of FLAN-T5.
Notably, on more challenging datasets, FM2 and \textsc{ParaGen}, there exists a relatively larger performance gap between FLAN-T5$_{\textsc{780m}}$ and FLAN-T5$_{\textsc{3b/11b}}$.

\section{Analysis on LLM-based Verifiers}
This section further analyzes the influence of given evidence on LLM-based verifiers~(\S\ref{evidence}), as well as their robustness~(\S\ref{robust}) and generalization ability~(\S\ref{generalization}). Appendix~\ref{memorization} also investigates how LLMs' memorization of inputs may influence their judgments. And Appendix~\ref{case_study_sec} further shows several representative cases that LLMs fail to judge, to provide more insights.

\subsection{Influence of Given Evidence}\label{evidence}
The aforementioned experiments reveal the importance of retrieving external knowledge for fact verification. We further assess how different types of  evidence influence performance. Tab.~\ref{retriever_analysis} shows: \textbf{(1)} Golden evidence is better than retrieved one,
both outperforming null and random evidence. \textbf{(2)} ChatGPT is more susceptible to irrelevant information in given evidence than FLAN-T5$_{\textsc{11b}}$. For example, ChatGPT is superior with solely golden evidence but underperforms with mixed golden and random evidence. This can potentially elucidate ChatGPT’s suboptimal performance when using retrieved evidence that may include irrelevant information. \textbf{(3)} ChatGPT performs much better than FLAN-T5$_{\textsc{11b}}$ if given relevant but fake~(Adv) or contradictory evidence~(Adv+Gold), which may be prevalent cases when retrieving evidence from the Internet~\cite{shu2017fake}. This reflects ChatGPT's more limited reliance on given evidence and stronger conviction in its internal knowledge. Nevertheless, the overall drop in accuracy compared to using golden evidence confirms the susceptibility of LLMs to false information~\cite{bian2023drop}. 


\subsection{Robustness}\label{robust}

\begin{table}[!t]
\small
\centering
\begin{adjustbox}{max width=\columnwidth}
\begin{tabular}{@{} lccm{0.001em}cc @{}}
\toprule
\multirow{2}{*}{{\textbf{Verifiers}}}&\multirow{2}{*}{{\textbf{Setting}}}&\multirow{2}{*}{{\textbf{\textsc{SentCom}}}}
&&\multicolumn{2}{c}{\textbf{\textsc{ParaGen}}}\\
\cline{5-6}
&&&&\textbf{Sent}
&{\textbf{Para}}\\
\midrule
\multirow{2}{*}{\textbf{FLAN-T5$_{\rm\textsc{11b}}$}} &ZS 
& \textbf{74.3}$_{0.8}$
&& \textbf{65.6}$_{1.7}$
& \textbf{77.7}$_{3.0}$ \\
 &FS
&\underline{73.3}$_{0.8}$
&& \underline{60.0}$_{0.8}$
& \underline{67.2}$_{2.0}$ \\
\hline
\multirow{2}{*}{\textbf{GPT3.5}} &ZS
 & 62.9$_{1.5}$
 && 44.0$_{5.7}$
 & 64.6$_{5.7}$ \\
 &FS
& 68.0$_{1.7}$
&& 52.9$_{6.0}$
& 71.7$_{4.4}$ \\
\hline
\multirow{2}{*}{\textbf{ChatGPT}}~&ZS
& \textcolor{black}{69.6$_{4.8}$}
 && \textcolor{black}{53.6$_{3.0}$}
 & \textcolor{black}{72.4$_{4.4}$} \\
&FS
& \textcolor{black}{70.2$_{1.3}$}
&& \textcolor{black}{51.1$_{2.2}$}
& \textcolor{black}{66.9$_{3.1}$} \\

\bottomrule

\end{tabular}
\end{adjustbox}
\caption{Pearson's correlations averaged across four prompts. The subscript indicates the standard deviation. ZS/FS means the zero/few-shot setting. Under the few-shot setting, we insert five manually selected demonstrations before each testing example.}
\label{robustness_stat}
\end{table}

LLMs are widely observed to be sensitive to synonymous prompts~\cite{jiang2020can}. 
We design three additional prompts by changing the question in Tab.~\ref{ipt_prompt} (details in Appendix~\ref{robust_prompt}).
Tab.~\ref{robustness_stat} shows the mean and standard deviation on \textsc{mgs} across four prompts. 
We find (1) FLAN-T5$_{\rm\textsc{11b}}$ and ChatGPT are better under the zero-shot setting, while GPT3.5 is better under the few-shot setting. 
(2) FLAN-T5$_{\rm\textsc{11b}}$ is more robust to different prompts than GPT3.5 and ChatGPT with lower variance. 

\subsection{Generalization Ability}\label{generalization}
It is crucial for fact verifiers to deal with various inputs~\cite{garbacea2019judge}. We focus on LLMs' generalization to statements generated by different models, depending on the context or not, or involving different types of named entities.

\begin{figure}[!t]
  \centering
\includegraphics[width=\linewidth]{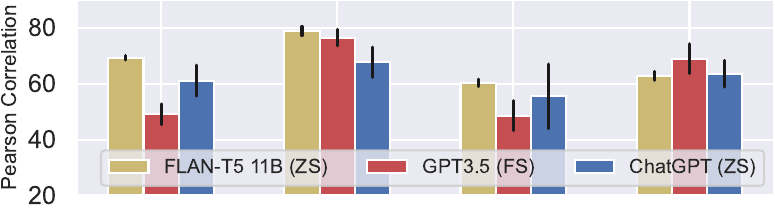}\\
\includegraphics[width=\linewidth]{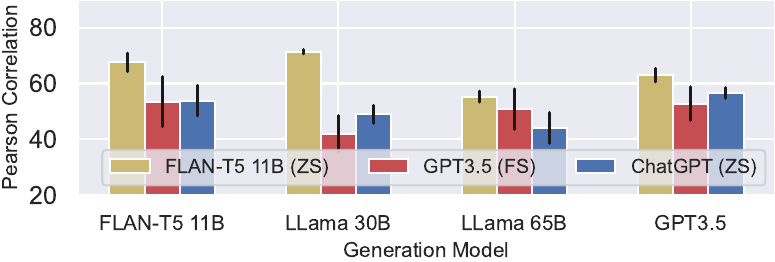}
  \caption{Pearson's correlations when verifying outputs from different generation models. \textbf{Top}: \textsc{SentCom}; \textbf{Bottom}: \textsc{ParaGen}~(Sent). We calculate the standard deviation~(black bars) across four instructions in \S\ref{robust}.}
  \label{fig:generalization}
\end{figure}

\noindent\textbf{Smaller models struggle to verify outputs from larger models.}
Fig.~\ref{fig:generalization} shows 
FLAN-T5$_{\rm\textsc{11b}}$ performs worse at verifying outputs from relatively large models~(LLama$_{\rm\textsc{65b}}$ and GPT3.5). ChatGPT is more stable across different generation models. 

\noindent\textbf{Verifying context-dependent statements 
is more challenging and can be enhanced through de-contextualization.}
For each sentence in \textsc{ParaGen}~(Sent), we manually annotate whether it refers to any entities in the context. 
Tab.~\ref{depend_stat} shows: \textbf{(1)} Performance on context-dependent sentences will notably drop if the context is unseen, indicating the verifiers are indeed utilizing the given context to judge the factuality. De-contextualization by first performing coreference resolution~(CR) can bring substantial improvement. \textbf{(2)} Performance on context-independent sentences becomes even better without context, suggesting the context may introduce noise. CR hardly further improves the performance. 
\textbf{(3)} De-contextualization at the sentence level also benefits paragraph-level verification.
Appendix~\ref{context_fact} shows more experiment details. 

\begin{table}[!t]
\centering
\begin{adjustbox}{max width=\columnwidth}
\begin{tabular}{@{} l ccc m{0.001em} ccc m{0.001em} c@{}}
\toprule
\multirow{3}{*}{\textbf{Verifiers}}&\multicolumn{3}{c}{\textbf{Sent}}&&\multicolumn{3}{c}{\textbf{Sent}}&&\multirow{2}{*}{\textbf{Para}}\\
&\multicolumn{3}{c}{\textbf{w/ Dependencies}}&&\multicolumn{3}{c}{\textbf{w/o Dependencies}}&\\
\cline{2-4}
\cline{6-8}
\cline{10-10}
&\textbf{AUR}&\textbf{AUP}&\textbf{$r$}&&\textbf{AUR}&\textbf{AUP}&\textbf{$r$}&&\textbf{$r$}\\
\midrule
\textbf{\textbf{FLAN-T5$_{\rm\textsc{11b}}$} (ZS)}
& \underline{87.4} & \underline{72.6} & \underline{59.9} 
&
& 88.6 & 84.7 & 69.1 && 79.4\\
~~\textbf{w/o Context}
& 81.0 & 58.5 & 49.5
&
& \underline{90.2} & \textbf{85.6} & \underline{72.3} &&\underline{82.9} \\
~~\textbf{w/ CR}
& \textbf{88.9} & \textbf{79.0} & \textbf{68.4}
&
& \textbf{90.9} & \underline{85.2} & \textbf{72.8} &&\textbf{86.1}\\
\midrule
\textbf{ChatGPT (ZS)}
& \textcolor{black}{\underline{75.1}} & \textcolor{black}{\underline{38.1}} & \textcolor{black}{\underline{41.1}}
&
& \textcolor{black}{84.1} & \textcolor{black}{62.9} & \textcolor{black}{63.4} &&68.1\\
\textbf{~~w/o Context}
& \textcolor{black}{72.2} & \textcolor{black}{35.0} & \textcolor{black}{36.5}
&
& \textcolor{black}{\underline{84.8}} & \textcolor{black}{\underline{65.9}} & \textcolor{black}{\underline{66.0}} & &\underline{74.4}\\
\textbf{~~w/ CR}
& \textcolor{black}{\textbf{79.6}} & \textcolor{black}{\textbf{45.7}} & \textcolor{black}{\textbf{52.0}}
&
& \textcolor{black}{\textbf{85.1}} & \textcolor{black}{\textbf{66.4}} & \textcolor{black}{\textbf{67.1}} && \textbf{78.1}\\

\bottomrule

\end{tabular}
\end{adjustbox}
\caption{Generalization to sentences with or without dependencies on the context. \textbf{w/o Context:} not providing the context during verification. \textbf{w/ CR:} first performing coreference resolution using ChatGPT to eliminate potential dependencies in a sentence and then verifying it without context.}
\label{depend_stat}
\end{table}

\begin{figure}[!t]
  \centering
\includegraphics[width=\linewidth]{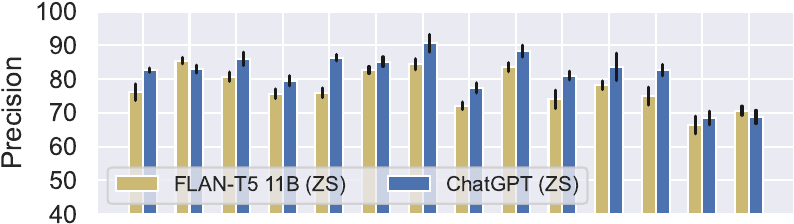}\\
\includegraphics[width=\linewidth]{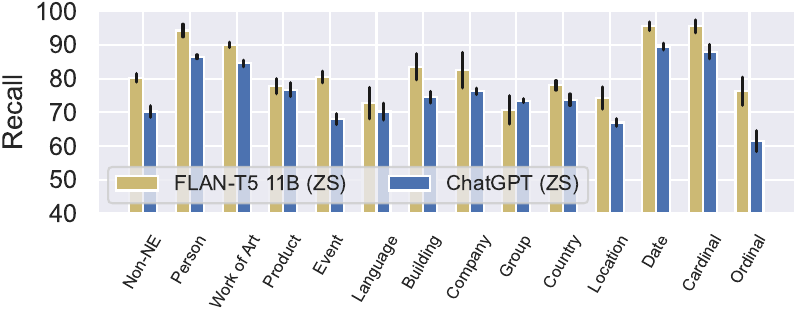}\\
  \caption{Precision and recall on the \texttt{unfactual} category~
  varying with entity types. ``Non-NE'' means non-named-entity common words~(e.g., ``computer'').}
  \label{fig:faviq}
\end{figure}

\noindent\textbf{Numeral-involved statements are more difficult to verify.} The factuality of a statement mainly manifests in the entities involved and inter-entity relations~\cite{nan2021entity}. We are curious whether the verifiers perform similarly at verifying statements concerning different types of entities. We experiment on the FaVIQ dataset~\cite{park2022faviq}, which consists of 188k sentences converted from information-seeking QA pairs. The original answer to the question is a word or phrase, and the resulting sentence is factual only if the answer is correct, enabling us to know which part of a sentence is unfactual. We categorize all sentences by the entity types\footnote{We use spaCy~\cite{spacy2} for named entity recognition.} of corresponding answers and randomly sample 100 factual and 100 unfactual sentences in each category to test the verifiers.

Fig.~\ref{fig:faviq} shows: 
\textbf{(1)} LLMs perform better at ``Person,'' ``Work of Art,'' and ``Building;'' 
\textbf{(2)} The performance is poor at numeral-related types ``Cardinal'' and ``Ordinal.'' 
Particularly, 
LLMs tend to misidentify factual sentences with cardinal numerals as unfactual. 
The reasons lie in the difficulty of  
reasoning over inter-numeral logical relations, and retrieving numerals-related evidence. For example, the evidence recall@10 score of the ``Cardinal'' type is $\sim$26\% lower than that of ``Person'' (0.43 vs. 0.58). More details are in Appendix~\ref{ent_type}.

\section{Conclusion}
We present a comprehensive study around two research questions: the extent to which current LLMs hallucinate; and their potential as effective fact verifiers. Firstly, we quantitatively affirm the significant hallucination issue of current LLMs through a well-designed human evaluation. This  highlights the urgent need for powerful fact verifiers to measure and incentivize progress in LLMs' factuality. To this end, we repurpose the LLMs into fact verifiers, and examine their ability to judge the factuality of model-generated and human-crafted statements. Further analysis shows their heavy reliance on high-quality evidence and discusses their robustness and generalization ability.
The evaluation suite and implemented verifiers in this paper can facilitate further research on fact verifiers and trustworthy generation models. 

\section*{Acknowledgements}
This work was supported by the National Key Research and Development Program of China (No.
2021ZD0113304), the National Science Foundation for Distinguished Young Scholars (with No.
62125604), and the NSFC projects (Key project with No. 61936010).
\section*{Limitations}
We summarize our limitations as follows: 
\paragraph{I. Regarding the Quantification of Hallucinations} 
\begin{compactitem}
    \item[\textbf{(a)}] We only focus on quantifying hallucinations for retrieval-free open-ended generation. Although \citet{shuster2021retrieval} showed that retrieval could significantly reduce hallucinations, LLMs are observed to overly depend on retrieved texts for generation~(e.g., directly copying from those texts) regardless of fluency and relatedness with input instructions~\cite{liu2023evaluating}, 
    which is out of our scope.
    \item[\textbf{(b)}] The model outputs are generated under the few-shot setting to ensure that they are as objective and informative as possible, although most generation models are used under the zero-shot setting in reality.     
\end{compactitem}

\paragraph{II. Regarding Evaluation Sets}
\begin{compactitem}
    \item[\textbf{(a)}] Examples in \textsc{mgs} are limited to the Wikipedia domain due to the difficulty of manually annotating statements involving much professional knowledge in other domains. To mitigate this issue, we include model-generated statements from the news domain (e.g., XsumFaith) in \textsc{dss} and conduct experiments on ChatGPT-generated statements in the question-answering domain (Appendix~\ref{factprompts}). we expect to collect examples in more professional domains~(e.g., finance) in future work. 
\item[\textbf{(b)}]Statements in all three evaluation sets rarely require multi-hop reasoning for judging, considering that outputs of current LLMs are organized as chains of single-hop ones.

\end{compactitem}
\paragraph{III. Regarding Fact Verifiers} 
\begin{compactitem}
    \item[\textbf{(a)}] Although we endeavor to evaluate LLMs' ability to deal with contradictory or fake evidence in \S\ref{evidence}, the auto-constructed evidence does not occur naturally and may exhibit many artifacts. Contradictory or fake evidence hardly occurs when retrieved from Wikipedia, but it will be much more common if retrieving evidence from the Internet.
    \item[\textbf{(b)}] We do not perfectly answer why FLAN-T5 is less susceptible to irrelevant information in retrieved evidence than ChatGPT due to the lack of knowledge about the implementation details of OpenAI's GPT series.    
    We conjecture that the reasons possibly lie in that the instruction-tuning data of FLAN-T5 models have included lots of noisy inputs, leading to better robustness to irrelevant information. In contrast, ChatGPT's better knowledge memorization ability may make it less influenced by the relevant but fake information in the evidence.
    The causal factors underlying these results remain to be investigated in the future work.
    \item[\textbf{(c)}] We only equip LLMs with single-step interaction with external corpora through the retriever. In future work, we expect to build autonomous reasoning agents that can verify any texts through multiple-step interaction with diverse knowledge environments~\cite{guan2024amor}.
\end{compactitem}




\section*{Ethics Statements}
Our experiment results are based on existing public resources~(datasets, model checkpoints, and codes). We use widely adopted settings for model generation and evaluation, making our analysis easily replicated. We resorted to Amazon Mechanical Turk (AMT) for human evaluations of model-generated statements. We did not ask about personal privacy or collect any personal information of annotators in the annotation process. We pay each worker \$2.5/\$7.5 for each HIT task of \textsc{SentCom}/\textsc{ParaGen}, respectively, leading to an hourly wage of $\sim$\$30, which is much higher than the minimum wage of \$7.5/h in the U.S. We decided the payment according to the average length of data examples. We admit that there may still be unpredictable bias in \textsc{mgs} even though we have carefully reviewed all annotated results from an ethical perspective. 


\bibliography{anthology,custom}
\appendix

\section{Quantifying LLMs' Hallucination}
\subsection{Generating Statements}\label{gen_statements}
The statements in \textsc{mgs} are generated under the few-shot setting with manually selected in-context demonstrations. For \textsc{SentCom}, the demonstrations are selected directly from the FEVER training set. And for \textsc{ParaGen}, the demonstrations are selected by sampling an entity from all Wikipedia titles as the input~(not overlapping the inputs for generation) and taking the first five sentences of the corresponding introduction section as the golden output. Tab.~\ref{demo_mgs} shows the demonstrations.
We ensure that all words in each input entity of demonstrations and generation outputs for \textsc{ParaGen} appear more than 100 times in the WebText corpus to avoid rare words. 

\subsection{Human Evaluation}\label{man_ant}
Fig.~\ref{fig:ant} in the main paper shows the annotation interface. To control the annotation quality, we discard those submissions where workers (1) do not provide evidence for sentences annotated as factual or unfactual; (2) do not annotate sentences in golden statements as factual; (3) annotate more than 90\% sentences as factual since manual inspection finds that the faction of factual sentences is much less than 90\%. 
When there are sentences assigned with different labels by three workers in a statement, we collect two additional annotations for it and retain three annotations with the highest agreement for each sentence. We obtain the final human judgments by repeating the above steps until each sentence has three valid annotations.

We ask workers to provide evidence mainly in order to force workers to search for related information to help them judge the factuality. Tab.~\ref{golden_mgs} shows that LLMs get better results using golden evidence than retrieved evidence on \textsc{mgs}, indicating the high quality of collected evidence.

\subsection{Human Judgements for No-Evidence Statements}\label{no_evidence_stat}
There are three ground-truth labels for each statement in \textsc{mgs}, i.e., \texttt{Factual}, \texttt{Unfactual}, and \texttt{No Evidence}. To convert human judgments to numerical scores, we regard \texttt{Factual} as 1 and both \texttt{Unfactual} and \texttt{No Evidence} as 0, following prior work that converted the 3-way classification to the 2-way classification~\cite{sarlin2020superglue,sathe2020automated}. We would still like to investigate whether LLMs perform differently to verify unfactual statements and no-evidence ones. As shown in Tab.~\ref{no_evidence}, the conclusion is similar under both settings: FLAN-T5$_{\textsc{11b}}$ outperforms ChatGPT under both settings. This indicates that it is reasonable to merge unfactual and no-evidence statements as one category.

\begin{table}[!t]
\centering
\begin{adjustbox}{max width=\columnwidth}
\begin{tabular}{@{}lccc m{0.001em} ccc@{}}
\toprule
\multirow{2}{*}{\textbf{Verifiers}}&\multicolumn{3}{c}{\textbf{\textsc{SentCom}}}
&&\multicolumn{3}{c}{\textbf{\textsc{ParaGen}~(Sent)}}\\
\cline{2-4}
\cline{6-8}
&\textbf{AUR}&\textbf{AUP}&\textbf{$r$}&&\textbf{AUR}&\textbf{AUP}&\textbf{$r$}\\
\midrule
\multicolumn{8}{c}{{\cellcolor[gray]{.95}}\textbf{Retrieving Evidence from External Corpora}}\\
\hline
\textbf{FLAN-T5$_{\textsc{11b}}$}&93.1&96.9&74.9&&88.0&80.9&66.8\\
\midrule
\multicolumn{8}{c}{{\cellcolor[gray]{.95}}\textbf{Using Golden Evidence}}\\
\hline
\textbf{FLAN-T5$_{\textsc{11b}}$}&96.0&98.2&80.6&&90.7&84.4&72.0\\
\midrule
\multicolumn{8}{c}{{\cellcolor[gray]{.95}}\textbf{Not Using any Evidence}}\\
\hline
\textbf{FLAN-T5$_{\textsc{11b}}$}&77.1&87.0&43.2&&70.9&49.7&32.5\\
\bottomrule
\end{tabular}
\end{adjustbox}
\caption{Results on \textsc{mgs} under different settings.} 
\label{golden_mgs}
\end{table}

\begin{table}[!t]
\small
\centering
\begin{adjustbox}{max width=\columnwidth}
\begin{tabular}{@{} lccc m{0.001em} ccc @{}}
\toprule
\multirow{2}{*}{\textbf{Verifiers}} & \multicolumn{3}{c}{\textbf{\textsc{SentCom}}} &&\multicolumn{3}{c}{\textbf{\textsc{ParaGen}~(Sent)}}\\
\cline{2-4}
\cline{6-8}
&\textbf{AUR}&\textbf{AUP}&\textbf{$r$}&&\textbf{AUR}&\textbf{AUP}&\textbf{$r$}\\
\midrule
\multicolumn{8}{c}{{\cellcolor[gray]{.95}}\textbf{Distinguishing Factual Statements from Unfactual Ones}}\\
\hline
\textbf{FLAN-T5$_{\textsc{11b}}$}&93.0&97.7&70.7&&90.2&88.3&71.4\\
\textbf{ChatGPT}&83.3&90.5&62.2&&82.8&66.4&63.6\\
\midrule
\multicolumn{8}{c}{{\cellcolor[gray]{.95}}\textbf{Distinguishing Factual Statements from No-Evidence Ones}}\\
\hline
\textbf{FLAN-T5$_{\textsc{11b}}$}&93.3&99.0&62.7&&84.7&87.2&62.1\\
\textbf{ChatGPT}&87.9&96.4&60.4&&75.0&65.7&52.3\\
\bottomrule
\end{tabular}
\end{adjustbox}
\caption{Results on \textsc{SentCom} and \textsc{ParaGen}~(Sent) to distinguish factual statements from unfactual/no-evidence ones. The number of \texttt{factual}/\texttt{unfactual}/\texttt{no-evidence} is 129/43/21 in \textsc{SentCom}, and is 155/247/166 in \textsc{ParaGen}~(Sent).}
\label{no_evidence}
\end{table}

\begin{table*}[!ht]
\small
\centering
\begin{adjustbox}{max width=\linewidth}
\begin{tabular}{@{}p{450pt}@{}}
\toprule
\textbf{\textsc{SentCom} demonstrations:}\\
1. \textcolor{red}{\textbf{Swedish Empire}} was ruled by Gustavus Adolphus from 1611 to 1632.\\
2. \textcolor{red}{\textbf{The Boston}} Celtics play their home games at TD Garden.\\
3. \textcolor{red}{\textbf{Chris Hemsworth}} appeared in A Perfect Getaway.\\
4. \textcolor{red}{\textbf{History of}} art includes architecture, dance, sculpture, music, painting, poetry literature, theatre, narrative, film, photography and graphic arts.\\
5. \textcolor{red}{\textbf{Nikolaj Coster-Waldau}} worked with the Fox Broadcasting Company.\\
\midrule
\textbf{\textsc{ParaGen} demonstrations:}\\
1. \textcolor{red}{\textbf{Fire and Darkness}} is a cancelled three-dimensional real-time strategy video game developed by Singularity Software. The game consists of a player controlling one of two factions, and their main mission is to defeat the enemy faction to secure the planet's resources. Its development started in 1996 and lasted for three years, with developers working mostly on summer. Although the project was incomplete, it became the first game to win the Seumas McNally Grand Prize at the Independent Games Festival of 1999. The development team invested time, but no money into the project.\\
2. \textcolor{red}{\textbf{Patrick Sharp}} (born December 27, 1981) is a Canadian former professional ice hockey player who played 15 seasons in the National Hockey League (NHL) for the Philadelphia Flyers, Chicago Blackhawks, and Dallas Stars. Sharp played collegiate hockey at the University of Vermont before he was drafted by the Flyers in 2001. He began his NHL career with the Flyers organization, but was traded to the Blackhawks in 2005. He became a three-time Stanley Cup champion with the Blackhawks in 2010, 2013, and 2015. Sharp was later dealt to the Stars in 2015, where he spent two seasons before returning to the Blackhawks in 2017.\\
3. The \textcolor{red}{\textbf{Cleveland East Ohio Gas Explosion}} occurred on the afternoon of Friday, October 20, 1944. The resulting gas leak, explosion and fires killed 130 people and destroyed a one square mile area on Cleveland, Ohio's east side. At 2:30 p.m. on the afternoon of Friday, October 20, 1944, above ground storage tank number 4, holding liquefied natural gas in the East Ohio Gas Company's tank farm, began to emit a vapor that poured from a seam on the side of the tank. The tank was located near Lake Erie on East 61st Street, and winds from the lake pushed the vapor into a mixed use section of Cleveland, where it dropped into the sewer lines via the catch basins located in the street gutters. As the gas mixture flowed and mixed with air and sewer gas, the mixture ignited.\\
4. \textcolor{red}{\textbf{Broke Sky}} is a 2007 neo-noir 35 millimeter film, and the directorial debut of cinematographer Thomas L. Callaway. The film stars Will Wallace, Joe Unger, Bruce Glover, Duane Whitaker and Barbara Chisholm, and has earned comparisons to the work of the Cohen Brothers. Bucky and Earl are the two man team that collect and dispose of road kill for the county. A new, specially designed carcass removal truck forces them to choose which one of them gets to keep his job and who is let go. Earl comes up with a plan so they can both keep their jobs, but it means working at night.\\
5. \textcolor{red}{\textbf{Design technology}}, or D.T., is the study, design, development, application, implementation, support and management of computer and non-computer based technologies for the express purpose of communicating product design intent and constructability. Design technology can be applied to the problems encountered in construction, operation and maintenance of a product. At times there is cross-over between D.T. and Information Technology, whereas I.T. is primarily focused on overall network infrastructure, hardware \& software requirements, and implementation, D.T.\\
\bottomrule
\end{tabular}
\end{adjustbox}
\caption{Demonstrations for generating statements in \textsc{mgs}. The \textcolor{red}{\textbf{red}} words correspond to \textcolor{red}{\{input\}} in Tab.~\ref{gen_prompt}.}
\label{demo_mgs}
\end{table*}


\section{Collecting \textsc{wks}}\label{collect_hcs}
\textsc{wks} aggregates seven existing fact verification datasets, including:
\begin{compactitem}
    \item[(1)] 
\textbf{FEVER:} It contains crowd-sourced statements by altering a word or negating 
sentences from Wikipedia~\cite{thorne-etal-2018-fever}, thereby leading to strong artifacts~\cite{schuster-etal-2019-towards}. We use the KILT version of FEVER~\cite{petroni2021kilt}. Since the official test set is hidden, we sample 1K examples from the validation set for testing and use the rest for validation. 
\item[(2)] \textbf{BoolQ-FV}~\cite{thorne2021evidence}: It consists of more realistic claims than FEVER, which are derived from real-world information needs 
by rewriting users' search-engine queries and verifying them against evidence from Wikipedia.
\item[(3)] \textbf{FM2}~\cite{eisenschlos2021fool}: It is collected by
gamification. Players write challenging statements supported or refuted by evidence from Wikipedia and spot refuted claims written by others.
\item[(4)] \textbf{PubMedQA}~\cite{jin2019pubmedqa}: It is initially a question-answering dataset specifically designed for the biomedical domain based on the PubMed database\footnote{\url{https://pubmed.ncbi.nlm.nih.gov/}}, which is a comprehensive collection of biomedical literature. Each PubMedQA example consists of a yes-no question and the abstract of the corresponding background paper. We prompt ChatGPT to convert the question into a declarative sentence as the statement to be judged with the abstract as the golden evidence. 
\item[(5)] \textbf{XsumFaith}~\cite{maynez2020faithfulness} and \textbf{SummEval}~\cite{fabbri2021summeval}: Each example consists of a summary and the corresponding source document. The summaries are generated by various models and paired with human-annotated binary faithfulness labels to the source documents. We regard the summary as a statement to be judged and the source document as the golden evidence.
\item[(6)] \textbf{SciFact}~\cite{wadden-etal-2020-fact}: It contains scientific claims against a corpus of scientific papers. The claims are re-formulated from naturally occurring citation sentences. We use the dataset to test the generalization of LLMs to more professional domains than Wikipedia.
\end{compactitem}
We do not include FaVIQ used in \S\ref{generalization} in \textsc{wks} since (1) most statements of FaVIQ are transformed from information-seeking QA pairs, which is similar to BoolQ-FV; and (2) FaVIQ focuses on token-level factual errors, which have been covered by FEVER. To make \textsc{wks} less redundant, we do not include FaVIQ in \textsc{wks}.

\section{Experiments}
\subsection{Evaluation Metrics}\label{evaluation_metrics}
We use multiple metrics to evaluate the verifiers, including ECE, ACC, AUR, AUP, and $r$, etc. These metrics are focusing on different aspects. \textbf{(1)} ECE is the weighted average absolute difference between metric scores and accuracy in each bin of $[0,1]$ as follows:
\begin{align*}
\text{ECE}=\sum_{b=1}^B\frac{n_b}N|\text{acc}(b)-\text{conf}(b)|,
\end{align*}
where $\text{acc(b)}$ is the accuracy, $\text{conf(b)}$ is the average metric scores~(confidence), and $n_b$ is the number of examples in $b$-th bin. We set $B=20$ in our experiments. A well-calibrated verifier can serve as an estimation of probabilities of making mistakes, which is more informative than only predicting ``factual'' or not. Note that lower ECE does not mean better discrimination ability. Supposing that a model always predicts a score of 0.5 on a perfectly balanced dataset, ECE will be 0, while the model is useless. \textbf{(2)} ACC is the fraction of examples that are correctly predicted, which means both human judgments and metric scores are either larger or smaller than 0.5. 
\textbf{(3)} AUR is one of the commonest metrics to evaluate binary classifiers. Although both ACC and AUR test the discrimination ability, AUR does not assume a pre-defined threshold, so it is more important on imbalanced datasets~(e.g., {BoolQ-FV} and \textsc{ParaGen}). \textbf{(4)} AUP is also widely used for the evaluation of imbalanced datasets. 
For comparison, AUR/AUP is more sensitive to the discrimination ability of verifiers on the \texttt{factual}/\texttt{non-factual} category. Therefore, AUP is a better metric when most statements are non-factual~(e.g., \textsc{ParaGen}). \textbf{(5)} All the above metrics require human judgments to be binary, while Pearson's correlation $r$ can be used between two continuous variables~(e.g., \textsc{ParaGen}~(Para)). In summary, we recommend comprehensively considering different metrics from multiple perspectives in future research and realistic application.


\subsection{Retriever}\label{retriever}
We compare BM25 and Contriever~\cite{izacard2022unsupervised} in terms of recall of golden evidence on \textsc{wks}, i.e., the ratio of tokens in golden evidence that also appear in the top 10 pieces of retrieved evidence. Tab.~\ref{eval_ret} shows the much higher recall scores of Contriever, so we use it in our experiments.

Furthermore, Tab.~\ref{retriever_analysis_num} shows the Pearson's correlation scores of FLAN-T5$_{\textsc{11b}}$ and ChatGPT varying with the number of retrieved passages (each passage contains ~150 tokens) on the FM2 dataset. We see that ChatGPT reaches saturation faster than FLAN-T5$_{\textsc{11b}}$ with increased retrieved passages, although ChatGPT supports a maximum length of 4,096 tokens. This indicates the length extrapolation ability of FLAN-T5 to some extent\footnote{ The encoders of FLAN-T5 models are trained with a maximum length of 512 tokens. Thanks to the relative positional encoding mechanism adopted by FLAN-T5 (all relative distances are truncated into 512), we can test the performance of FLAN-T5 with much longer inputs.}. We agree that the training-testing length bias may potentially impact the performance of FLAN-T5, although we have not explicitly observed such an impact. Actually, when the length of retrieved passages is much larger than 512 in our paper, FLAN-T5$_{\textsc{11b}}$ always performs the best. We will further investigate the influence of the length bias in our future work.


\begin{table}[!t]
\centering
\begin{adjustbox}{max width=\columnwidth}
\begin{tabular}{l|cccc}
\toprule
\textbf{Retrievers}&\textbf{FEVER}&\textbf{BoolQ-FV}&\textbf{FM2}&\textbf{SciFact}\\
\midrule
\textbf{BM25}&88.09&63.88&75.50&55.71\\
\textbf{Contriever}&94.95&88.63&87.80&79.00\\
\bottomrule
\end{tabular}
\end{adjustbox}
\caption{Token recall scores~(\%) of different retrievers.}
\label{eval_ret}
\end{table}

\begin{table}[!t]
\centering
\begin{adjustbox}{max width=\columnwidth}
\begin{tabular}{@{} lccccc @{}}
\toprule
\textbf{\# Retrieved Passages} & \textbf{1} & \textbf{3} & \textbf{5} & \textbf{7} & \textbf{10} \\
\midrule
\textbf{FLAN-T5$_{\textsc{11b}}$} &54.2	&64.0&	67.4&	68.4&	68.8\\
\textbf{ChatGPT}&	47.9	&58.2&	58.4&	57.9&	56.7\\
\bottomrule
\end{tabular}
\end{adjustbox}
\caption{Pearson's correlation on FM2 varying with the number of retrieved passages on the verification performance.}
\label{retriever_analysis_num}
\end{table}

\subsection{Verifiers}
\paragraph{NLI$_{\textsc{11b}}$} The corresponding model card on HuggingFace is ``google/t5\_xxl\_true\_nli\_mixture''. It is trained on a mixture of SNLI~\cite{bowman2015large}, MNLI~\cite{williams-etal-2018-broad}, FEVER~\cite{thorne-etal-2018-fever}, Scitail~\cite{khot2018scitail}, PAWS~\cite{zhang2019paws} and
VitaminC~\cite{schuster2021get}. Although FEVER is included in the training set of the model, we still regard this metric as unsupervised since they use golden evidence for training while we mainly use retrieved evidence for testing.
\paragraph{FiD} Tab.~\ref{acc_fid} shows the dataset transfer results of the supervised verifier FiD on \textsc{wks}, illustrating the poor generalization ability of supervised verifiers with a significant drop in accuracy when transferring it to a different dataset. On the other hand, when applying FiD on \textsc{ParaGen}~(Sent), we do not provide the context during verification since FiD has not been trained to utilize the context.

\begin{table}[!t]
\centering
\begin{adjustbox}{max width=\columnwidth}
\begin{tabular}{@{} l|cccc @{}}
\toprule
\multirow{2}{*}{\textbf{Training Sets}}&\multicolumn{4}{c}{\textbf{Test Sets}}\\
&\textbf{FEVER}&\textbf{BoolQ-FV}&\textbf{FM2}&\textbf{SciFact}\\
\midrule
\textbf{FEVER}&\textbf{94.60}&77.65&70.00&75.92\\
\textbf{BoolQ-FV}&83.10&\textbf{82.71}&61.30&70.68\\
\textbf{FM2}&89.40&74.88&\textbf{76.96}&70.16\\
\textbf{SciFact}&65.80&71.94&52.61&\textbf{78.53}\\
\bottomrule
\end{tabular}
\end{adjustbox}
\caption{Accuracy of the FiD metric which are trained on one dataset and then used for another one.}
\label{acc_fid}
\end{table}

\paragraph{FLAN-T5} FLNA-T5 series have been trained on SciFact with golden evidence. Therefore, the results of FLAN-T5 on SciFact in Tab.~\ref{human_eval_stat_other} may be over-estimated. 
We ensure that FLAN-T5 series have not been trained on the other three datasets of \textsc{wks} in the Wikipedia domain, including FEVER, Boolq-FV, and FM2. 


\begin{table}[!t]
\small
\centering
\begin{adjustbox}{max width=\columnwidth}
\begin{tabular}{@{} lccc m{0.001em} ccc m{0.001em} ccc@{}}
\toprule
\multirow{2}{*}{\textbf{Datasets}} & \multicolumn{3}{c}{\textbf{FEVER}}&&\multicolumn{3}{c}{\textbf{BoolQ-FV}} &&\multicolumn{3}{c}{\textbf{FM2}}\\
\cline{2-4}
\cline{6-8}
\cline{10-12}
&\textbf{ECE}&\textbf{AUR}&\textbf{$r$}&&\textbf{ECE}&\textbf{AUR}&\textbf{$r$}&&\textbf{ECE}&\textbf{AUR}&\textbf{$r$}\\
\midrule
\textbf{Soft}&\textbf{3.1}&\textbf{98.2}&\textbf{90.2}&&\textbf{10.2}&\textbf{94.7}&\textbf{75.3}&&\textbf{8.2}&\textbf{89.5}&\textbf{68.8}\\
\textbf{Hard}&6.2&93.9&87.7&&14.5&87.6&70.0&&18.0&81.9&65.0\\
\bottomrule
\end{tabular}
\end{adjustbox}
\caption{Results of hard coding and soft coding based on FLAN-T5$_{\textsc{11b}}$.}
\label{hard_soft}
\end{table}

\paragraph{ChatGPT}\label{chatgpt} We hard code the predicted factuality score of ChatGPT as 1/0 if it returns ``A''/``B'' using the prompt in Tab.~\ref{ipt_prompt} because its output probabilities are unavailable. Tab.~\ref{hard_soft} compares the results of hard coding~(i.e., using the \emph{generated answers}) and soft coding~(i.e., using the \emph{generation probabilities}) based on FLAN-T5$_{\textsc{11b}}$, indicating that the performance of ChatGPT may be underestimated in terms of metrics that depend on probabilities,  including ECE, AUR, AUP, and $r$ in our experiments.

\begin{table}[!t]
\centering
\begin{adjustbox}{max width=\columnwidth}
\begin{tabular}{@{}lccc m{0.001em} ccc}
\toprule
\multirow{2}{*}{\textbf{Verifiers}}&\multicolumn{3}{c}{\textbf{FEVER}}
&&\multicolumn{3}{c}{\textbf{FM2}}\\
\cline{2-4}
\cline{6-8}
&\textbf{ECE}&\textbf{ACC}&\textbf{AUR}&&\textbf{ECE}&\textbf{ACC}&\textbf{AUR}\\
\midrule
\textbf{Constant}&48.3&51.7&50.0
&&50.6&49.4&50.0\\
\midrule
\multicolumn{8}{c}{{\cellcolor[gray]{.95}}\textbf{Retrieving Evidence from External Corpora}}\\
\hline
\textbf{LLama$_\textsc{7b}$}&45.9&51.6&47.8
&&47.7&49.4&45.0\\
\textbf{Alpaca$_\textsc{7b}$}&43.9&51.3&56.3
&&45.3&49.3&46.2\\
\hline
\textbf{FLAN-T5$_{\textsc{11b}}$}&\textbf{3.1}&\textbf{93.8}&\textbf{98.2}
&&\textbf{8.2}&\textbf{82.0}&\textbf{89.5}\\
\hline
\textbf{InstructGPT}&\underline{6.4}&88.8&96.1
&&\underline{15.0}&71.5&81.5\\
\textbf{GPT3.5}&7.6&91.7&\underline{96.6}
&&21.0&77.5&\underline{82.8}\\
\textbf{ChatGPT}&8.2&\underline{92.8}&92.6
&&21.4&\underline{79.1}&78.6\\
\midrule
\multicolumn{8}{c}{{\cellcolor[gray]{.95}}\textbf{Using Golden Evidence}}\\
\hline
\textbf{LLama$_\textsc{7b}$}&6.3&59.6&64.8
&&\textbf{6.4}&54.6&55.8\\
\textbf{Alpaca$_\textsc{7b}$}&43.8&52.1&86.4
&&46.7&49.2&73.9\\
\hline
\textbf{FLAN-T5$_{\textsc{11b}}$}&\underline{5.4}&93.3&\textbf{98.6}
&&\underline{9.7}&\underline{87.3}&\textbf{96.1}\\
\hline
\textbf{InstructGPT}&\textbf{3.9}&\underline{93.9}&\underline{98.0}
&&8.0&85.4&92.6\\

\textbf{GPT3.5}&6.1&93.8&96.6
&&12.1&87.1&93.2\\
\textbf{ChatGPT}&5.8&\textbf{94.4}&94.3
&&11.6&\textbf{88.5}&\underline{88.4}\\
\midrule
\multicolumn{8}{c}{{\cellcolor[gray]{.95}}\textbf{Not Using any Evidence}}\\
\hline
\textbf{LLama$_\textsc{7b}$}&14.7&48.5&67.5
&&\textbf{6.7}&50.5&49.4\\
\textbf{Alpaca$_\textsc{7b}$}&\textbf{5.4}&74.9&80.5
&&19.9&51.8&54.4\\
\hline
\textbf{FLAN-T5$_{\textsc{11b}}$}&\underline{11.1}&74.4&\textbf{87.0}
&&\underline{15.7}&\underline{59.5}&\underline{66.3}\\
\hline
\textbf{InstructGPT}&14.4&\underline{75.3}&83.4
&&26.4&58.4&62.9\\
\textbf{GPT3.5}&25.6&73.8&78.3
&&41.4&57.3&61.5\\
\textbf{ChatGPT}&18.1&\textbf{82.0}&\underline{81.5}
&&33.5&\textbf{66.7}&\textbf{66.5}\\

\bottomrule
\end{tabular}
\end{adjustbox}
\caption{Performance of different instruction-tuned LLMs on FEVER and FM2. All models are under the zero-shot setting.} 
\label{other_model}
\end{table}

\paragraph{More LLMs}\label{more_llms} In addition to FLAN-T5 series, GPT3.5 and ChatGPT, we also test the performance of LLama$_{\textsc{7b}}$, Alpaca$_{\textsc{7b}}$~\cite{alpaca} and InstructGPT on \textsc{wks}. Tab.~\ref{other_model} shows the results under different settings. We observe that:
\begin{compactitem}
    \item[\textbf{(1)}]  InstructGPT is better calibrated than GPT3.5 with similar discrimination performance under all settings. Compared with CPT3.5, InstructGPT has not been trained through RLHF~(reinforcement learning from human feedback), indicating that RLHF may impact calibration, which is accordant with GPT4's technical report~\cite{openai2023gpt4}. 
    \item[\textbf{(2)}] LLama$_{\textsc{7b}}$ performs like random guessing on FM2 under all settings. When using retrieved evidence, its performance is also at the random level on FEVER, which can be improved using golden evidence, suggesting its poor robustness to noise in the given evidence. Surprisingly, when not using any evidence, the AUR score of LLama$_{\textsc{7b}}$ on FEVER instead increases compared with using golden evidence, despite the worse ECE and ACC scores. This means LLama$_{\textsc{7b}}$ is seriously biased toward a certain category when not using evidence, and providing golden evidence can alleviate the bias but impact the discrimination ability.
    \item[\textbf{(3)}] Alpaca$_{\textsc{7b}}$ is comparable to or better than LLama$_{\textsc{7b}}$ on the whole. One contrary phenomenon to LLama$_{\textsc{7b}}$ is that Alpaca$_{\textsc{7b}}$ is better calibrated with a higher ACC score when not using any evidence than using golden evidence on FEVER, indicating that the instruction tuning fashion of Alpaca$_{\textsc{7b}}$ increases the bias toward a certain category when using golden evidence, despite a better ability to utilize internal and external knowledge.
\end{compactitem}

\begin{table}[!t]
\small
\centering
\begin{adjustbox}{max width=\columnwidth}
\begin{tabular}{@{}p{220pt}@{}}
\toprule
The task is to judge whether the statements are factually correct based on the corresponding facts. Two examples are as follows:\\
Example 1. \\
Facts:\\
0. Baseball Hall of Fame in 1988 $\cdots$\\
1. He was a carpenter and construction worker by trade. $\cdots$\\
2. players, and 599 of them loved Willie Stargell $\cdots$\\\\
Statement: Willie Stargell lived with his father after his mother and father separated.\\\\
Output: Willie Stargell lived with his mother in Florida after his parents' divorce, and later returned to live with her in California. There is no mention of him living with his father after the separation. So the statement is incorrect; the answer is No.\\\\\\
Example 2.\\
Facts:\\
0. Alfredo Stroessner Alfredo Stroessner Matiauda $\cdots$\\
1. in most other Latin American countries $\cdots$\\
2. led to the liberalization of Paraguay $\cdots$\\
3. system. On 3 February 1989, Stroessner was $\cdots$\\
4. some of Stroessner's policies$\cdots$\\\\
Statement: After the overthrow of Stroessner, Paraguay had its first non-military head of state in four decades.\\\\
Output: Stroessner, a military officer, had been in power for 35 years, from 1954 to 1989, and had been overthrown in a military coup led by General Andrés Rodríguez. After Stroessner's ousting, Rodríguez orchestrated a political campaign with the Colorado Party and won the presidency in a multi-party election held in May 1989. This marked the first time in four decades that Paraguay had a non-military head of state. So the statement is correct; the answer is Yes.\\\\\\

Now please judge the following statement:\\
\midrule
The task is to rate the factuality level for the statement based on the fraction of details in the statement that can be supported by the facts.\\
You should output only one single number from 1, 2, 3, 4, and 5, indicating the factuality score of the statement, where a higher score means that more details of the statement are factual.\\\\
1 - Almost all details in the statement are not supported by the facts.\\
2 - Some details in the statement are supported by the facts, but the majority are not.\\
3 - About half of the details in the statement are supported by the facts.\\
4 - Most of the details in the statement are supported by the facts.\\
5 - All details in the statement are supported by the facts.\\\\
Please output the score of the following statement:\\
\bottomrule
\end{tabular}
\end{adjustbox}
\caption{Input Prompt for CoT prompting~(Top) and Likert-scale rating~(Bottom).}
\label{likert_scale_prompt}
\end{table}

\begin{table}[!t]
\centering
\begin{adjustbox}{max width=\columnwidth}
\begin{tabular}{@{}lcccc m{0.001em} ccccc@{}}
\toprule
\multirow{2}{*}{\textbf{Methods}}&\multicolumn{4}{c}{\textbf{FEVER}}
&&\multicolumn{4}{c}{\textbf{FM2}}\\
\cline{2-5}
\cline{7-10}
&\textbf{ACC}&\textbf{P}&\textbf{R}&\textbf{$r$}&&\textbf{ACC}&\textbf{P}&\textbf{R}&\textbf{$r$}\\
\midrule
\textbf{Direct}&\textbf{92.8}&\underline{92.8}&\underline{92.8}&\textbf{92.8}&&\textbf{79.1}&\underline{80.1}&\underline{75.2}&\textbf{56.7}\\
\textbf{CoT}&87.5&\textbf{93.2}&{81.8}&75.7&&70.7&\textbf{83.1}&51.1&44.5\\
\textbf{Likert-Scale}&\underline{89.6}&86.9&\textbf{93.4}&\underline{79.6}&&\underline{76.1}&74.6&\textbf{76.1}&\underline{53.5}\\
\bottomrule
\end{tabular}
\end{adjustbox}
\caption{Performance of different verification methods. \textbf{Direct} means directly predicting ``A'' or ``B'' as described in \S\ref{verification_method}.}
\label{cot_likert}
\end{table}

\begin{table}[!t]
\small
\centering
\begin{adjustbox}{max width=\columnwidth}
\begin{tabular}{@{} lccc @{}}
\toprule
\textbf{Models}&\textbf{ACC}&\textbf{AUR}&\textbf{$r$}\\
\midrule
\multicolumn{4}{c}{{\cellcolor[gray]{.95}}\textbf{Retrieving Evidence from External Corpora}}\\
\midrule
\textbf{FLAN-T5$_{\rm\textsc{11b}}$} & 64.4 & \textbf{68.1} & \textbf{26.5}\\
\textbf{ChatGPT}&\textbf{71.7} & 59.4 & 19.5\\
\midrule
\multicolumn{4}{c}{{\cellcolor[gray]{.95}}\textbf{Not Using any Evidence}}\\
\midrule
\textbf{FLAN-T5$_{\rm\textsc{11b}}$} & 63.5 & \textbf{67.6} & \textbf{26.0}\\
\textbf{ChatGPT}&\textbf{72.1} & 50.5 & 1.6\\
\bottomrule
\end{tabular}
\end{adjustbox}
\caption{Results on FactPrompts.}
\label{fact_pro}
\end{table}

\paragraph{Results on FactPrompts}\label{factprompts} The dataset is released by a contemporary work~\cite{chern2023factool}, which comprises real-world prompts from various sources, such as Quora and TruthfulQA~\cite{lin2022truthfulqa}, along with corresponding responses generated by ChatGPT. We regard the response as the statement to be judged and retrieve 10 passages from C4~\cite{raffel2020exploring} using an off-the-shelf tool\footnote{\url{https://c4-search.apps.allenai.org/}} as the evidence. FactPrompts include 177 factual statements and 56 unfactual ones. All these statements are model-generated, so we believe that they should be unseen during the training stages of LLMs. Nevertheless, we acknowledge that texts with similar distributions to these statements might still be seen.

As shown in Tab. \ref{fact_pro}, FLAN-T5$_{\rm\textsc{11b}}$ outperforms ChatGPT with or without external evidence. ChatGPT without any evidence has a random performance on FactPrompts (AUR$\approx$50.0). This might be because the statements in the dataset are generated by ChatGPT itself, so ChatGPT easily regards them as factual statements.

\subsection{Verification Method}\label{verification_method2}
Besides using generation probabilities of certain answer words~(e.g., ``A,'' and ``B'') as factuality scores described in \S\ref{verification_method}, we also try several other methods widely used for NLG evaluation as follows:
\begin{compactitem}
    \item \textbf{Chain-of-Thought Prompting~(CoT)} was proposed to elicit LLMs' reasoning ability~\cite{weichain}, and recently was also used for improving NLG evaluators~\cite{liu2023gpteval}. 
    \item \textbf{Likert-Scale Rating} is the most common method for human and automatic evaluation~\cite{gao2023human}. We prompt LLMs to rate statements from 1 to 5 and re-scale the ratings into the range from 0 to 1 as the final factuality scores.
\end{compactitem}
Tab.~\ref{likert_scale_prompt} shows the prompt to instruct LLMs to judge statements using the above two verification methods. We compare different verification methods based on ChatGPT and show the results in Tab.~\ref{cot_likert}:
\begin{compactitem}
    \item[\textbf{(1)}] Directly predicting ``A'' or ``B'' is the simplest but best verification method.
    \item[\textbf{(2)}] Using CoT prompting will lead to a significant performance drop. We attribute the drop to the impact of generated non-factual rationales, which misleads ChatGPT to mistake factual statements for non-factual ones, as indicated by the similar precision but lower recall than direct prediction.
    \item[\textbf{(3)}] When using Likert-scale rating, LLMs tend to give higher scores than direct prediction, with a lower precision but higher recall. The reason may be that the ground-truth labels of the statements are collected more extremely than Likert-scale rating, which means only very minor factual errors can make a statement labeled \texttt{unfactual}. For example, a statement with a ground-truth label of 0 can be rated 4~(0.75 after re-scaling) under the Likert-scale rating. Therefore, Likert-scale rating may also be a good choice for fact verification if users expect the score to be less strict and more informative. In future research, we recommend focusing on standardizing Likert-scale protocols for factuality rating and designing efficient and effective methods for detecting fine-grained factual errors~(e.g., at the token level).
\end{compactitem}

\subsection{Prompts for Robustness Assessment}\label{robust_prompt}
We design different prompts to assess the robustness of LLMs in \S\ref{robust} by changing the ``Question'' part in Tab.~\ref{ipt_prompt} to:
\begin{compactitem}
    \item ``Is the statement entailed by the given facts? (A) Yes. (B) No. Please answer A or B:''
    \item ``Based on the given facts, judge whether the statement is factually correct. Please answer Yes or No:''
    \item ``Can the given facts support the statement? Please answer Yes or No:''
\end{compactitem}

\subsection{Influence of Non-Factual Context}\label{context_fact}
\begin{table}[!t]
\small
\centering
\begin{adjustbox}{max width=\columnwidth}
\begin{tabular}{@{}lccc@{}}
\toprule
\textbf{Proportion (\%)}&\textbf{Factual}&	\textbf{Unfactual}&\textbf{No Evidence}\\
\midrule
\textbf{Factual Context}&45.0&45.5&9.5\\
\textbf{Non-Factual Context}&15.1&42.1&42.7\\
\bottomrule
\end{tabular}
\end{adjustbox}
\caption{Label distribution of sentences in \textsc{ParaGen} (Sent) with factual or non-factual context. Here, factual context means the context of the sentence includes only factual sentences, and non-factual context means the opposite.} 
\label{nonfact_context}
\end{table}

\paragraph{Influence on Model Generation}
In Tab.~\ref{fig:prop}, we observe the increase of no-evidence sentences as the generation proceeds to later sentences. We conjecture that it is because of the error accumulation during generation. A contemporary work~\cite{zhang2023language} also highlights the phenomenon and empirically attributes it to error propagation. We show our finding in Tab.~\ref{nonfact_context}: There is a lower proportion of no-evidence sentences when the context is factual than non-factual. Here, factual context means the context of the sentence includes only factual sentences, and non-factual context means the opposite.

Intuitively, more errors in the context (unfactual or no-evidence) correlate to more no-evidence sentences. However, it may be difficult to affirm the causality between them. The potential insight is that when generating open-ended long texts using auto-regressive models, it is necessary to dynamically involve external feedback, e.g., from humans, retrievers, planners, etc., to avoid the impact of non-factual contexts.

\paragraph{Dependency Annotation} For each sentence in \textsc{ParaGen} (Sent), we manually annotate whether it involves dependencies on the context, i.e., including references to any entities in the context.
We label a sentence \texttt{independent} if 
\begin{compactitem}
    \item[\textbf{(1)}] It is the first one in the paragraph~(137 sentences totally).
    \item[\textbf{(2)}] It does not contain nouns or pronouns that refer specifically to entities in the context.
    \item[\textbf{(3)}] It contains nouns that refer to entities in the context but can be understood solely. For example, in the paragraph {``Russian wine is produced in several regions. $\cdots$ Russian wines have won numerous international awards.''}, the second ``Russian wine'' in the last sentence refers to the same entity as the context, but the last sentence is not dependent on the context.
    \end{compactitem}
Otherwise, the sentence is labeled \texttt{dependent} on the context. We asked two graduates for annotation and found they annotate the same labels for more than 90\% sentences. Finally, we obtain 189 context-dependent sentences~(61 factual and 128 non-factual), and 279 context-independent ones~(94 factual and 185 non-factual).

\paragraph{De-Contextualization} We use ChatGPT for coreference resolution~(CR) using the prompt in Tab.~\ref{cr_prompt}. To assess the accuracy of the CR method, we inspect twenty randomly sampled CR results and find ChatGPT can complete the task perfectly without any errors. This may be because all sentences in a paragraph of \textsc{ParaGen} are almost centered on the same entity so that it is easy to resolve coreferences. 

\begin{table}[!t]
\small
\centering
\begin{adjustbox}{max width=\columnwidth}
\begin{tabular}{@{} p{220pt} @{}}
\toprule
Task: Given a statement following its context, find all coreferences to the context in the statement, and replace them with words that they refer to. Be careful not to change the original word order as much as possible.\\
For example, if the context is ``Mary loves pizzas.'', and the statement following the context is ``She eats them every day.'', you should output ``Mary eats pizzas every day.''\\
Now finish the following task:\\
Context: \textcolor{red}{\{$c$\}}\\
Statement following the context: \textcolor{red}{\{$s$\}}\\
Output:\\
\bottomrule
\end{tabular}
\end{adjustbox}
\caption{Prompts to instruct ChatGPT to perform coreference resolution.}
\label{cr_prompt}
\end{table}

\begin{table}[!t]
\centering
\begin{adjustbox}{max width=\columnwidth}
\begin{tabular}{@{} l ccc m{0.001em} ccc @{}}
\toprule
\multirow{2}{*}{\textbf{Verifiers}}& \multicolumn{3}{c}{\textbf{Factual Cont~(104/127)}} && \multicolumn{3}{c}{\textbf{Non-Factual Cont~(51/286)}}\\
\cline{2-4}
\cline{6-8}
& \textbf{AUR}&\textbf{AUP}&\textbf{$r$} && \textbf{AUR}&\textbf{AUP}&\textbf{$r$}\\
\midrule
\textbf{\textbf{FLAN-T5$_{\rm\textsc{11b}}$} (ZS)}
& 91.6
& 91.8
& 74.2
&
& 84.1
& 49.8
& 46.4\\
\textbf{~~w/ CR}
& 93.1 
& 92.5 
& 76.0
&
& 85.8
& 62.6
& 58.4\\
\midrule
\textbf{ChatGPT (ZS)}
& 83.3
& 72.4
& 66.1
&
& 76.3
& 30.5
& 38.3\\
\textbf{~~w/ CR}
& 82.1
& 71.9
& 64.0
&
& 82.5
& 40.4
& 52.1\\
\bottomrule
\end{tabular}
\end{adjustbox}
\caption{Generalization to sentences whose context is factual or non-factual. \textbf{Cont} is short for \textbf{context}. Two digits in each parenthesis are the number of factual/non-factual statements. \textbf{Factual Context} means all sentences in the context of a statement are factual, or the context is empty. \textbf{Non-Factual Context} means there exists at least one non-factual sentence in the context of the statement.}
\label{depend_factual_stat}
\end{table}

\begin{table*}[!t]
\centering
\begin{adjustbox}{max width=\textwidth}
\begin{tabular}{lllr@{.}c}
\toprule
\textbf{Entity Types}&\textbf{spaCy Label}&\textbf{Description}&\multicolumn{2}{l}{\textbf{$\frac34$ Percentile}}\\
\midrule
\textbf{Non-NE}&N/A&Common words that are not named entities&9,526&5\\
\textbf{Person}&PERSON& People, including fictional&420&0\\
\textbf{Work of Art}& WORK\_OF\_ART& Titles of books, songs, etc.&2,166&0\\
\textbf{Product}& PRODUCT& Objects, vehicles, foods, etc. (not services)&3,456&5\\
\textbf{Event}& EVENT& Named hurricanes, battles, wars, sports events, etc.&8,559&0\\
\textbf{Language}& LANGUAGE& Any named language&245,337&0\\
\textbf{Building}& FAC& Buildings, airports, highways, bridges, etc.&603&0\\
\textbf{Company}& ORG& Companies, agencies, institutions, etc.&18,557&5\\
\textbf{Group}& NORP& Nationalities or religious or political groups&21,983&5\\
\textbf{Country}& GPE& Countries, cities, states&205,839&0\\
\textbf{Location}& LOC& Non-GPE locations, mountain ranges, bodies of water&8,624&75\\
\textbf{Date}& DATE& Absolute or relative dates or periods&372,927&0\\
\textbf{Cardinal}& CARDINAL& Numerals that do not fall under another type&3,572,566&0\\
\textbf{Ordinal}& ORDINAL& ``first'', ``second'', etc.&50,076&0\\
\bottomrule
\end{tabular}
\end{adjustbox}
\caption{Entity Types and corresponding labels and descriptions from spaCy. $\frac34$ percentile means $\frac34$ of entities of some entity type appear in fewer Wikipedia passages than that number.}
\label{spacy_label}
\end{table*}

\begin{table}[!t]
\centering
\begin{adjustbox}{max width=\columnwidth}
\begin{tabular}{@{}lrc@{}}
\toprule
\multicolumn{3}{c}{\textbf{Correlation between Perplexity and Predicted Factuality Score}}\\
\midrule
\textbf{Models}	&\textbf{FEVER}&	\textbf{FM2}\\
\midrule
FLAN-T5$_{\textsc{780m}}$&-16.4**&-11.6**\\
FLAN-T5$_{\textsc{3B}}$&-11.0**&-15.2**\\
FLAN-T5$_{\textsc{11B}}$&-10.3**&-12.2**\\
\bottomrule
\toprule
\multicolumn{3}{c}{\textbf{Correlation between Perplexity and Golden Factuality Score}}\\
\midrule
\textbf{Models}	&\textbf{FEVER}&	\textbf{FM2}\\
\midrule
FLAN-T5$_{\textsc{780m}}$&-5.7**&3.4**\\
FLAN-T5$_{\textsc{3B}}$&-2.6**&2.5**\\
FLAN-T5$_{\textsc{11B}}$&-3.6**&3.7**\\
\bottomrule
\end{tabular}
\end{adjustbox}
\caption{Correlation scores between FLAN-T5's perplexity and predicted/golden factuality scores on FEVER and FM2. ** means p-value$<$0.01.}
\label{mem_corr}
\end{table}


\begin{table*}[!t]
\small
\centering
\begin{adjustbox}{max width=\textwidth}
\begin{tabular}{@{} cp{100pt} p{140pt}ccp{108pt} @{}}
\toprule
\textbf{ID}&\textbf{Statement}&\textbf{Retrieved Most-Related Evidence}&\textbf{Label}&\textbf{Pred}&\textbf{ChatGPT Explanation}\\
\midrule
1&Georges Bizet died in his early {\textbf{40s}}.\newline \textcolor{red}{(\textit{from} FM2)}&\faCheck~\textcolor{blue}{(Georges Bizet)} Georges Bizet (25 October 1838 – 3 June 1875), registered at birth as Alexandre César Léopold Bizet, was a French composer of the Romantic era.&0&1&Georges Bizet was born in 1838 and died in 1875, which means he was 36 years old when he died. This is considered to be in his early 40s. \\
\midrule
2&A function is continuous when it is differentiable at that point.\newline\textcolor{red}{(\textit{from} \textsc{BoolQ-FV})}&\faCheck~\textcolor{blue}{(Differentiable function)} If f is differentiable at a point x0, then f must also be continuous at x0. In particular, any differentiable function must be continuous at every point in its domain.&1&0&
A function can be continuous at a point but not differentiable at that point.\\
\midrule
3&\textbf{{The United States}} was the allied nation that was the first to reach the German capital of Berlin.\newline\textcolor{red}{(\textit{from} FaVIQ)}&\faTimes~\textcolor{blue}{(2nd Armored Division (United States))} After a rest in February, the division, now commanded by Major General Isaac D. White, drove on across the Rhine on 27 March, and was the first American division to reach the Elbe at Schonebeck on 11 April.&0&1&The U.S. Army's 2nd Armored Division was the first American division to reach the Elbe at Schonebeck on 11 April and entered Berlin in July, making the statement correct.\\
\midrule
4&Institute for Employment Studies is a \textbf{government-linked} think tank on \textbf{British} employment policy and practice. \newline\textcolor{red}{(\textit{from} \textsc{ParaGen})}&\faCheck~\textcolor{blue}{(Institute for Employment Studies)} The Institute for Employment Studies (IES) is a British centre of research and consultancy on human resources and employment.&0&1&The Institute for Employment Studies is a British think tank that focuses on research and consultancy related to human resources and employment.\\
\bottomrule
\end{tabular}
\end{adjustbox}
\caption{Several representative cases where both FLAN-T5$_{\textsc{11b}}$ and ChatGPT makes mistakes. We highlight non-factual words in the statements in \textbf{bold} and indicate the \textcolor{red}{source datasets} after the statements. ``Label'' or ``Pred'' refer to ground-truth labels or model predictions, and 1 or 0 means factual or not.  ChatGPT explanations are generated by prompting ChatGPT to explain its predictions. \textbf{Most-related evidence} means the evidence that is the most related to the ChatGPT explanation, which is manually selected from the top ten pieces of evidence from the retriever. For the evidence, we indicate the corresponding \textcolor{blue}{Wikipedia page titles} before them and use \faCheck and \faTimes~to illustrate whether the evidence can help judge the factuality of the statements.}
\label{case_study}
\end{table*}

\begin{table}[!t]
\small
\centering
\begin{adjustbox}{max width=\columnwidth}
\begin{tabular}{@{}p{220pt}@{}}
\toprule
\textbf{Statement:} Pope Gregory XIII was a \textcolor{blue}{{math teacher}} before becoming pope.\\
\midrule
\textbf{Evidence: } $\cdots$ 
after \textcolor{red}{studying mathematics, Latin, law}, and philosophy, initially at the university of Bologna, and then in Padua in 1545 and becoming a notary. His interests successively extended to \textcolor{red}{philosophy and logic, which he combined with the study of medicine}. He \textcolor{red}{obtained a degree in medicine and philosophy} $\cdots$ He \textcolor{red}{later taught jurisprudence for some years} $\cdots$\\
\midrule
\textbf{ChatGPT Explanation:} The facts state that Pope Gregory XIII \textcolor{red}{studied mathematics} and graduated in law, later \textcolor{red}{teaching jurisprudence for some years}. He also studied \textcolor{red}{philosophy and logic, which he combined with the study of medicine, obtaining a degree in medicine and philosophy}. Therefore, it can be inferred that \textcolor{cyan}{{he had a background in mathematics before becoming pope.}}\\
\bottomrule
\end{tabular}
\end{adjustbox}
\caption{An unfactual statement that ChatGPT fails to judge while FLAN-T5$_{\rm\textsc{11b}}$ succeeds. We obtain the explanation by prompting ChatGPT to explain its prediction. \textcolor{red}{{Red}} words are copied from the evidence, and \textcolor{cyan}{{cyan}} and \textcolor{blue}{{blue}} words are conflicting.} 
\label{case_chat}
\end{table}

\paragraph{Influence on Fact Verification} Tab.~\ref{depend_stat} in the main paper has shown greater difficulty in verifying context-dependent statements than context-independent ones. 
We are also curious whether non-factual context will impact the performance of the fact verifiers. 
Tab.~\ref{depend_factual_stat} answers the question in the affirmative. Performing CR will alleviate the disparity by improving the performance on statements with non-factual context. 

\subsection{Entity Type}\label{ent_type}
Tab.~\ref{spacy_label} shows details of entity types used in Fig.~\ref{fig:faviq}. We see that ``Person,'' ``Work of Art,'' and ``Building'' have relatively lower $\frac34$ percentile. This means these types of entities are distributed more sharply, which may account for better performance on the  statements corresponding to these types than other types.

\subsection{Influence of LLMs' Memorization}\label{memorization}
We are curious about whether there is a degree of memorization when LLMs make judgments about factuality. For example,  they may inherently tend to judge those texts that are more probable under them as hallucinatory regardless of the given evidence. To investigate this correlation, we take perplexity as a proxy to measure the memorization degree of an LLM for any one statement. Then, we compute Pearson’s correlation between the LLM’s perplexity and predicted/golden factuality score on the FEVER and FM2 test sets. It is intractable to compute the perplexity under GPT models since their APIs do not return the logits of any specified texts. Therefore, we only compute the correlation based on the FLAN-T5 models, and use its decoder to compute the perplexity. Tab.~\ref{mem_corr} shows the correlation results.

We see that the perplexity of FLAN-T5 correlates significantly with the predicted factuality score: When the LLM memorizes a statement better (indicated by a lower perplexity score), it is more likely to regard the statement as factual. However, the perplexity score does not correlate with the golden label. The results suggest that memorization can influence LLMs’ judgments about factuality to some extent, but may not be the main contributor to the superior fact verification performance of LLMs.

\begin{figure}[!t]
  \centering
\includegraphics[width=\linewidth]{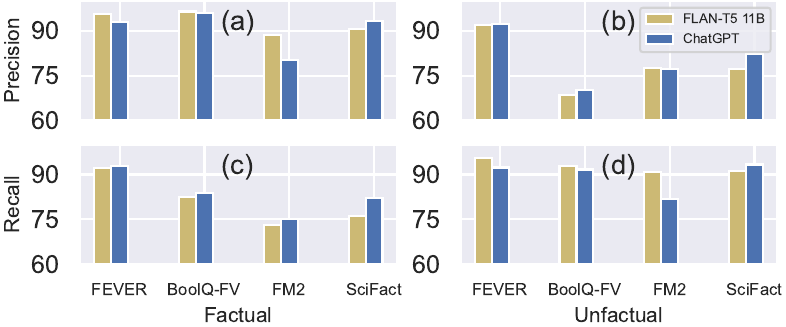}\\
  \caption{Precision and recall scores with retrieved evidence in \textsc{wks}~(\textbf{Left:} Factual \textbf{Right:} Unfactual). \textbf{Takeaway:} ChatGPT prefers to predict \texttt{factual}.}
  \label{fig:pr}
\end{figure}

\subsection{Case Study}\label{case_study_sec}
Fig.~\ref{fig:pr} plots precision and recall to better understand the preference of different LLMs. 
On SciFact, ChatGPT surpasses FLAN-T5$_{\textsc{11b}}$ 
with higher precision and recall. 
In the Wikipedia domain, ChatGPT exhibits a preference for predicting \texttt{factual} in contrast to FLAN-T5$_{\textsc{11b}}$, resulting in higher recall on factual statements~(Fig. \ref{fig:pr} c). Such a preference also leads ChatGPT to make more mistakes when predicting \texttt{factual}~(Fig. \ref{fig:pr} a) and misidentify unfactual statements as factual ones~(Fig. \ref{fig:pr} d).
Manual inspection reveals that ChatGPT easily builds spurious connections between related yet distinct concepts, as indicated in the following case study. 
On the other hand, \citet{zhong2023can} also emphasized the inadequacy of ChatGPT in assessing inter-sentence similarity compared with BERT-based models, which further supports our observation.

Tab.~\ref{case_chat} shows a case that ChatGPT fails to judge while FLAN-T5$_{\textsc{11b}}$ does not, suggesting that ChatGPT may build spurious connections between related yet distinct concepts~(e.g., ``math teacher'' and ``study mathematics'').

Through manual inspection of test examples that both ChatGPT and FLAN-T5$_{\textsc{11b}}$ make wrong judgments, we summarize several typical types of errors. To illustrate these error types, Tab.~\ref{case_study} shows several representative cases: 
\begin{compactitem}
    \item[\textbf{(1)}] \textbf{False Numerical and Logical Reasoning:} In Example 1 and 2, despite correct retrieval, ChatGPT mistakes ``36 years old'' for ``40s'', and uses ``Continuity is not sufficient for differentiability'' to refute ``differentiability is sufficient for continuity,'' indicating the weakness of LLMs in understanding numerical and logical relations. \item[\textbf{(2)}] \textbf{Misled Retriever:} In Example 3, the retriever is misled by the non-factual information~(i.e., ``the United States'') in the statement, thus leading to useless evidence and wrong predictions. When giving correct evidence to LLMs~(i.e., the Wikipedia page ``Race to Berlin''), we find they can make the right prediction. 
    \item[\textbf{(3)}] \textbf{Specious Inter-Entity Relations:} Statements generated by larger models~(such as GPT3.5) tend to seem more fluent and factual with interrelated entities, although the relations between them may be non-factual. In Example 4, the evidence provides useful information about ``IES,'' but it does not mention whether ``IES'' is linked to the government or its research is only limited to the British. ChatGPT ignores such specious relations and hence makes wrong predictions.
\end{compactitem}

\end{document}